\documentclass{article}

\usepackage{microtype}
\usepackage{graphicx}
\usepackage{subfigure}
\usepackage{booktabs} 

\usepackage[ruled,vlined,linesnumbered]{algorithm2e}

\usepackage{hyperref}

\usepackage[accepted]{icml2025}

\usepackage{amsmath}
\usepackage{amssymb}
\usepackage{mathtools}
\usepackage{amsthm}
\usepackage{amsmath, amssymb}

\usepackage[capitalize,noabbrev]{cleveref}

\theoremstyle{plain}
\newtheorem{theorem}{Theorem}[section]
\newtheorem{proposition}[theorem]{Proposition}
\newtheorem{lemma}[theorem]{Lemma}
\newtheorem{corollary}[theorem]{Corollary}
\theoremstyle{definition}
\newtheorem{definition}[theorem]{Definition}
\newtheorem{assumption}[theorem]{Assumption}
\theoremstyle{remark}

\usepackage[textsize=tiny]{todonotes}

\usepackage{pifont}

\usepackage[table,xcdraw]{xcolor}
\usepackage{multirow}
\usepackage{adjustbox}
\usepackage{subcaption}

\icmltitlerunning{Submission and Formatting Instructions for ICML 2025}

\begin{document}

\twocolumn[
\icmltitle{DANCE: Dynamic, Available, Neighbor-gated Condensation for Federated Text-Attributed Graphs}

\icmlsetsymbol{equal}{*}

\begin{icmlauthorlist}
\icmlauthor{Zekai Chen}{equal,yyy}
\icmlauthor{Haodong Lu}{equal,yyy}
\icmlauthor{Xunkai Li}{yyy}
\icmlauthor{Henan Sun}{sch}
\icmlauthor{Jia Li}{sch}
\icmlauthor{Hongchao Qin}{yyy}
\icmlauthor{Rong-Hua Li}{yyy}
\icmlauthor{Guoren Wang}{yyy}
\end{icmlauthorlist}

\icmlaffiliation{yyy}{Department of Computer Science, Beijing Institute of Technology, Beijing, China}
\icmlaffiliation{sch}{The Hong Kong University of Science and Technology (GZ), Guangzhou, China}
\icmlcorrespondingauthor{Rong-Hua Li}{lironghuabit@126.com}
\icmlkeywords{Machine Learning, ICML}

\vskip 0.3in
]



\printAffiliationsAndNotice{\icmlEqualContribution} 

\begin{abstract}

Federated graph learning (FGL) enables collaborative training on graph data across multiple clients. With the rise of large language models (LLMs), textual attributes in FGL graphs are gaining attention. Text-attributed graph federated learning (TAG-FGL) improves FGL by explicitly leveraging LLMs to process and integrate these textual features.
However, current TAG-FGL methods face three main challenges:
\textbf{(1) Overhead.} LLMs for processing long texts incur high token and computation costs. To make TAG-FGL practical, we introduce graph condensation (GC) to reduce computation load, but this choice also brings new issues.
\textbf{(2) Suboptimal.}
To reduce LLM overhead, we introduce GC into TAG-FGL by compressing multi-hop texts/neighborhoods into a condensed core with fixed LLM surrogates.
However, this one-shot condensation is often not client-adaptive, leading to suboptimal performance.
\textbf{(3) Interpretability.}
LLM-based condensation further introduces a black-box bottleneck: summaries lack faithful attribution and clear grounding to specific source spans, making local inspection and auditing difficult.
To address the above issues, we propose \textbf{DANCE}, a new TAG-FGL paradigm with GC. To improve \textbf{suboptimal} performance, DANCE performs round-wise, model-in-the-loop condensation refresh using the latest global model. To enhance \textbf{interpretability}, DANCE preserves provenance by storing locally inspectable evidence packs that trace predictions to selected neighbors and source text spans.
Across 8 TAG datasets, DANCE improves accuracy by \textbf{2.33\%} at an \textbf{8\%} condensation ratio, with \textbf{33.42\%} fewer tokens than baselines.

\end{abstract}

\section{Introduction}
Federated graph learning (FGL) enables collaborative training on graph data across multiple clients while preserving privacy by avoiding the sharing of raw data. With the rise of large language models (LLMs), the role of textual attributes in graph data has gained increasing attention~\cite{recommend_system,qiu2023app_gnn_fina3,qu2023app_gnn_bio2}. Text-attributed graph federated learning (TAG-FGL) improves the FGL framework by explicitly leveraging LLMs to process and integrate these textual features, enabling the learning process on text-rich graphs. This integration enhances FGL’s ability to leverage not only the graph’s structure but also the rich semantic information contained in node texts.

Recent approaches to TAG-FGL integrate language models in different ways, e.g., via shared global modules or federated backbones~\citep{wu2025fedbook,zhu2025fedgfm}, or by client-side LLM-based augmentation before standard subgraph-FL optimization~\citep{yan2025llm4fgl}. 
However, \emph{practical} TAG-FGL still faces three challenges.
\textbf{(1) Overhead.} 
Invoking LLMs to process long node texts incurs substantial token and computation costs, which can slow optimization under tight compute budget~\citep{yan2025llm4fgl,wu2025fedbook,zhu2025fedgfm}. 
To mitigate this overhead, we introduce \textbf{graph condensation (GC)} into TAG-FGL, aiming to reduce the amount of text and neighborhood information that must be processed.
\textbf{(2) Suboptimal.} 
A naive integration of GC often compresses multi-hop texts/neighborhoods into a small condensed core using \emph{fixed} LLM surrogates (summaries/embeddings). 
Such one-shot condensation is typically not tailored to each client’s graph and text distribution, causing client-specific representation mismatch and thus suboptimal performance.
\textbf{(3) Interpretability.} 
LLM-based condensation can further create a black-box bottleneck: the generated summaries/embeddings usually lack verifiable provenance and clear grounding to specific source spans, making it difficult to locally inspect and audit the decision process in regulated settings~\citep{lopez2024interplay,saifullah2024privacy}.

To address the above issues, we propose \textbf{DANCE} (\textbf{D}ynamic, \textbf{A}vailable, \textbf{N}eighbor-gated \textbf{C}ond\textbf{E}nsation), a new TAG-FGL paradigm built on \textbf{GC}.
To improve \textbf{suboptimal} performance caused by one-shot condensation, DANCE performs round-wise condensation refresh using the latest global model.
To enhance \textbf{interpretability}, DANCE preserves provenance by keeping locally inspectable evidence packs that trace predictions back to selected neighbors and source text spans.
Concretely, at each round, each client executes three steps in sequence:
(1) \emph{Label-aware node condensation} selects a condensed core as the candidate node set for GC;
(2) \emph{Hierarchical text condensation} selects evidence-providing neighbors and distills a compact set of readable summaries/evidence spans, while recording the selected neighbors and source spans as local evidence packs;
(3) \emph{Self-expressive topology reconstruction} rebuilds a lightweight propagation graph on the condensed core using fused graph--text representations to restore connectivity for downstream training.

\textbf{Contributions.}
\underline{\textbf{\textit{(1) New Formulation.}}} We introduce GC for TAG-FGL, which integrates interpretable LLM-based summary generation, dynamically adjusted to the evolving global model.
\underline{\textbf{\textit{(2) New Framework.}}} We present \textbf{DANCE}, a novel TAG-FGL framework that combines label-aware node condensation, hierarchical text condensation, and self-expressive topology reconstruction, enhancing both performance and interpretability.
\underline{\textbf{\textit{(3) SOTA Performance.}}} On 8 TAG datasets, DANCE achieves SOTA performance, surpassing competitive TAG-FGL baselines by 2.33\% on accuracy while reducing token processing costs by 33.42\% per condensed node.

\section{Preliminaries}
\label{sec:prelim}

\textbf{Text-Attributed Graphs.}
A text-attributed graph (TAG) is $G=(V,A,S)$, where $V$ is the node set,
$A\in\{0,1\}^{\lvert V\rvert\times\lvert V\rvert}$ is the adjacency matrix, and
$S=\{s_v\}_{v\in V}$ denotes node texts.
A text encoder maps each $s_v$ to a $d$-dimensional embedding:
\begin{equation}
\label{eq:text_encoder}
x_v=\mathrm{Enc}_{\psi}(s_v)\in\mathbb{R}^d,
\;
X=[x_v]_{v\in V}\in\mathbb{R}^{\lvert V\rvert\times d}.
\end{equation}
We study semi-supervised node classification, where a labeled subset $V_L\subset V$ is associated with labels $\{y_v\}_{v\in V_L}$.

\textbf{Graph Condensation for TAGs.}
Graph condensation constructs a compact surrogate TAG $\hat{G}$ that preserves task-relevant information of $G$ while reducing cost.
A condensed TAG is $\hat{G}=(\hat{V},\hat{A},\hat{S})$, where $\lvert\hat{V}\rvert\ll \lvert V\rvert$,
$\hat{S}=\{\hat{s}_v\}_{v\in\hat{V}}$ denotes condensed node texts, and
$\hat{A}\in\mathbb{R}^{\lvert\hat{V}\rvert\times\lvert\hat{V}\rvert}$ is a learned/reconstructed adjacency matrix (typically weighted and sparse).
We train the downstream predictor on the condensed TAG and aim to retain comparable node classification performance to training on the original TAG.

\textbf{Federated Learning on TAGs.}
The global TAG is distributed across $M$ clients; client $m$ holds a local subgraph $G^{(m)}=(V^{(m)},A^{(m)},S^{(m)})$.
At round $t$, the server broadcasts global model $\omega^{(t-1)}$ to clients $\mathcal{M}_t$. Each client optimizes locally and uploads an update $\Delta\omega_m^{(t)}$; the server aggregates
\vspace{-6pt}
\begin{equation}
\label{eq:federated_update}
\omega^{(t)}=\mathrm{Agg}\Big(\{\Delta\omega_m^{(t)}\}_{m\in\mathcal{M}_t}\Big),
\vspace{-6pt}
\end{equation}    
where $\mathrm{Agg}$ is an aggregation method.
Our goal is to learn global model $\omega$ for semi-supervised node classification while keeping raw texts/graphs local in the FL setting under communication constraints.
\section{Related Work}
\textbf{Graph neural networks.}
Graph neural networks (GNNs) learn node representations via neighborhood propagation and aggregation and are standard backbones for semi-supervised node classification~\citep{kipf2017semi,hamilton2017inductive,velickovic2018graph}. Later variants improve robustness on heterophily and structural heterogeneity via geometric encodings and heterophily-aware designs~\citep{pei2020geomgcn,zhu2020gprgnn,chen2020h2gcn}. In our setting, these models mainly serve as structural encoders that must be coupled with rich text attributes.

\textbf{Federated graph learning.}
FGL extends federated learning to graph-structured data, typically under a subgraph-FL protocol where each client holds a local subgraph and collaboratively trains a global GNN without sharing raw graphs~\citep{fu2022fgml,he2021fedgraphnn, wu2025comprehensive}. Prior work addresses non-IIDness via structural repair, personalization, and topology-aware aggregation, e.g., missing-neighbor generation~\citep{zhang2021subgraph}, personalized heads~\citep{baek2023personalized}, topology-aware confidence~\citep{li2024fedgta}, and condensation-based collaboration~\citep{fedc4}. However, most methods assume compact node features; when attributes are long texts, \emph{what} evidence to encode/propagate (and under what budgets) becomes a first-order bottleneck that is less explored.

\begin{figure*}[!t]
  \centering
  \includegraphics[width=\textwidth]{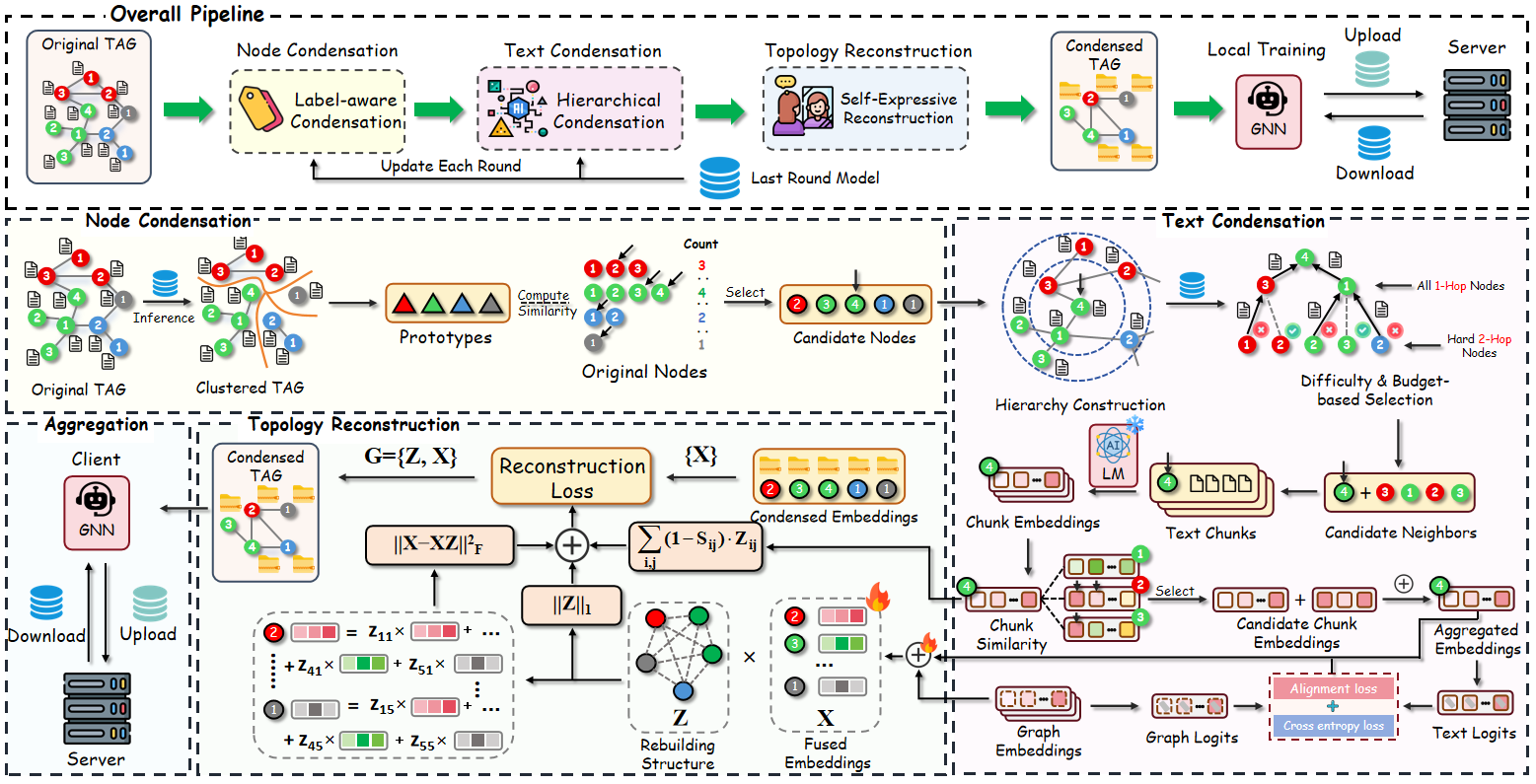}
  \caption{Overview of the proposed \textbf{DANCE} framework. Each client condenses a local TAG by (1) label-aware node condensation and (2) budgeted hierarchical text propagation/condensation, then (3) rebuilds a lightweight topology via self-expressive reconstruction on fused graph--text embeddings and (4) trains a GNN for federated aggregation.}
  \label{fig:framework}
\end{figure*}

\textbf{Federated learning on TAGs.}
As FGL enters text-rich domains, existing TAG-FGL methods incorporate language signals at different layers. (1) \emph{Client-side augmentation} uses LLMs or other generators to enrich local texts/neighborhoods before running standard subgraph-FL~\citep{yan2025llm4fgl}. (2) \emph{Federated modular coordination} shares global modules (e.g., codebooks/prompts/prototypes) to align topology--text learning without transmitting raw texts~\citep{wu2025fedbook,zhu2025fedgfm}. (3) \emph{Semantic-prior injection} regularizes message passing using label/ontology semantics or external knowledge~\citep{ji2025fassgat}. Despite these advances, most pipelines either treat text as largely static embeddings or absorb it into a monolithic backbone, while \emph{budgeted, round-adaptive} control of neighbor usage and chunk-level evidence retention is not a primary design focus. DANCE complements these efforts by making evidence selection and neighbor usage first-class, round-adaptive components under explicit budgets and federated privacy constraints, while keeping human-readable traces locally available.

\section{Methodology}
\label{sec:method}

\subsection{Overview}
In this section, we present a detailed description of \textbf{DANCE}. We introduce its three core modules: (1) Label-aware node condensation, (2) Hierarchical text condensation, and (3) Self-expressive topology reconstruction. We then summarize the complete round-wise client--server workflow in Alg.~\ref{alg:dance}, with an overview of the framework shown in Fig.~\ref{fig:framework}.

\subsection{Label-aware Node Condensation}
\label{sec:node_condense}

\textbf{Motivation.}
TAG-FGL is often bottlenecked by long-text processing and multi-hop propagation.
We therefore perform label-aware clustering-based node condensation to reduce computation and communication overhead.
We perform node condensation every 10 communication rounds.

\textbf{Model-conditioned pseudo-labels.}
At round $t$, client $m$ runs the last-round global model $\omega^{(t-1)}$ once on its local TAG to obtain node embeddings $\{z_v\}_{v\in V^{(m)}}$ and predicted labels $\hat{y}_v$, and their confidence scores $\pi_v$ (reusing cached text embeddings when available).
We assign pseudo-labels as follows:
\begin{equation}
\label{eq:pseudo_label}
\tilde{y}_v=
\begin{cases}
y_v, & v\in V_L^{(m)},\\
\hat{y}_v, & v\notin V_L^{(m)} \ \text{and}\ \pi_v\ge\tau,\\
\text{unassigned}, & v\notin V_L^{(m)} \ \text{and}\ \pi_v<\tau.
\end{cases}
\end{equation}
We use $\tau$ to keep only high-confidence pseudo-labels ($\pi_v\ge\tau$) when estimating class statistics. If a class becomes empty after filtering, we assign it zero budget and redistribute the remaining budget across the other classes.

\textbf{Distribution-preserving core selection.}
To capture intra-class diversity under client heterogeneity, we summarize each class $c$ by $P_c$ prototypes and score each node by its best prototype match:
\begin{equation}
\label{eq:proto_score}
\begin{aligned}
\mathcal{P}_c &= \mathrm{Cluster}\!\left(\{z_v\}_{v\in V_c^{(m)}}; P_c\right),
s_v = \max_{p\in \mathcal{P}_{\tilde{y}_v}} \kappa(z_v,p),
\end{aligned}
\end{equation}
where $P_c=\min(P,|V_c^{(m)}|)$, $\mathrm{Cluster}(\cdot)$ denotes mini-batch $k$-means, and $\kappa(\cdot,\cdot)$ is cosine similarity.
Given a compression ratio $r\in(0,1]$, we set the node budget $K=\lceil r|V^{(m)}|\rceil$ and perform label-stratified selection by allocating per-class quotas proportional to the confident pseudo-label distribution.
We then form the condensed core $\hat{V}^{(m)}_t$ by selecting the top nodes within each class according to $s_v$ until its quota is filled.
This yields a compact core that approximately preserves local label proportions while retaining representative nodes in the embedding space.
\subsection{Hierarchical Text Condensation}
\label{sec:text_condense}

\textbf{Motivation.}
Even on the condensed core $\hat{V}^{(m)}_t$, aggregating all multi-hop texts is costly and often redundant.
We thus perform budgeted evidence aggregation: (i) select a small set of evidence-providing neighbors under per-hop budgets, and (ii) distill a small set of text chunks under a token budget to form a compact evidence embedding for each core node.

\textbf{Frozen text bank and caching.}
For each node $u\in V^{(m)}$, we split its node text into short chunks
$\mathcal{S}(s_u)=\{s_{u,r}\}_{r=1}^{R_u}$, encode each chunk with a frozen text encoder
$e_{u,r}=\mathrm{Enc}(s_{u,r})\in\mathbb{R}^d$, and pool chunk embeddings to obtain a node-level text embedding
$t_u=\mathrm{Pool}(\{e_{u,r}\}_{r=1}^{R_u})\in\mathbb{R}^d$.
Here, $\mathrm{Pool}(\cdot)$ denotes softmax-based attention pooling.
Since $\mathrm{Enc}$ is fixed, $\{t_u\}$ can be cached across rounds.
At round $t$, we obtain graph-side embeddings $\{g_v\}$ by a forward pass of the last-round graph encoder on the \emph{original} local TAG.

\textbf{Budgeted neighbor gating.}
For a core node $v\in\hat{V}^{(m)}_t$, let $\mathcal{N}_v^{(\ell)}$ be its $\ell$-hop neighbors in the original local graph ($\ell\in\{0,1,2\}$).
We score candidates in $\tilde{\mathcal{N}}_v^{(\ell)}$($\tilde{\mathcal{N}}_v^{(0)}=\{v\}$) and select a sparse subset under hop budgets $B_\ell$ (Eq.~\eqref{eq:neighbor_score}--\eqref{eq:neighbor_gating}).
Given the significantly larger size of the 2-hop neighborhood, we additionally pre-filter $\mathcal N_v^{(2)}$ with difficulty score $u_v$ (Eq.~\eqref{eq:difficulty}).

We score a candidate neighbor $u\in\tilde{\mathcal{N}}_v^{(\ell)}$ by cross-modal attention score between $g_v$ and $t_u$:
\begin{equation}
\label{eq:neighbor_score}
s_{v,u}=\frac{(W_q g_v)^\top (W_k t_u)}{\sqrt{d}}.
\end{equation}
For each hop $\ell\in\{0,1,2\}$, we transform $\{s_{v,u}\}_{u\in\tilde{\mathcal{N}}_v^{(\ell)}}$ into hop-wise sparse attention weights $\alpha^{(\ell)}_{v,\cdot}$ under budget $B_\ell$:
\begin{equation}
\label{eq:neighbor_gating}
\alpha^{(\ell)}_{v,\cdot}
=
\Pi_{B_\ell}\!\Big(
\mathrm{entmax}\big(\{s_{v,u}\}_{u\in \tilde{\mathcal{N}}_v^{(\ell)}}\big)
\Big),
\end{equation}
where $\mathrm{entmax}(\cdot)$ yields sparse attention weights, selecting at most $B_\ell$ neighbors (implemented with a straight-through estimator).
We denote the selected nodes at hop $\ell$ as
$\mathcal{S}_v^{(\ell)}=\{u\in\tilde{\mathcal{N}}_v^{(\ell)}:\alpha^{(\ell)}_{v,u}>0\}$.
Then we aggregate the selected multi-hop evidence into a hierarchical context vector $c_v$:

\begin{equation}
\label{eq:hier_context}
c_v={\textstyle\sum\limits_{\ell\in\{0,1,2\}}}\gamma_\ell
      {\textstyle\sum\limits_{u\in\mathcal{S}_v^{(\ell)}}}\alpha^{(\ell)}_{v,u}\,t_u,\ 
\gamma_\ell\ge0,\ {\textstyle\sum\limits_{\ell}}\gamma_\ell=1.
\end{equation}

\textbf{Chunk selection.}
Neighbor gating determines \emph{which nodes} can provide evidence; we also select \emph{which chunks} to retain in the token budget $B_{\mathrm{tok}}$.
We collect chunk candidates only from the selected neighbors and denote the resulting candidate set as $\mathcal{E}_v$.Specifically, we form an attention query from the core-node graph embedding, $q_v=W_s g_v$, compute chunk-level attention scores $a_{v,(u,r)}$, and obtain budgeted sparse weights $\pi_{v,\cdot}$ via $\Pi_{B_{\mathrm{tok}}}(\cdot)$:

\begin{equation}
\label{eq:chunk_similarity}
\begin{gathered}
a_{v,(u,r)}=\frac{q_v^\top e_{u,r}}{\sqrt{d}},\\
\pi_{v,\cdot}=\Pi_{B_{\mathrm{tok}}}\!\Big(\mathrm{entmax}(\{a_{v,(u,r)}\}_{(u,r)\in\mathcal{E}_v})\Big),
\end{gathered}
\end{equation}

where $\mathrm{entmax}(\cdot)$ yields sparse attention weights, selecting at most $B_{tok}$ chunks (implemented with a straight-through estimator).
We aggregate selected chunks into a compact evidence embedding:

\begin{equation}
\label{eq:agg_text}
\tilde{t}_v=\sum_{(u,r)\in\mathcal{E}_v}\pi_{v,(u,r)}\,e_{u,r}.
\end{equation}

By construction, each core node uses at most $\sum_\ell B_\ell$ neighbors and at most $B_{\mathrm{tok}}$ chunks.
Gradients do not backpropagate into the frozen encoder $\mathrm{Enc}$; we update only the graph backbone and selection parameters (e.g., $W_q,W_k,W_s,\gamma$).

\textbf{Local evidence traces and summarization.}
For each core node, we store the top gated neighbors (via $\alpha^{(\ell)}_{v,\cdot}$) and selected evidence chunks (via $\pi_{v,\cdot}$), which can be mapped back to the original texts.
We then prompt an on-device LLM to summarize the selected chunks into short evidence summaries, which can be cached and reused across rounds. The prompt template is provided in Appendix~\ref{sec:prompt}. Neither raw texts, audit traces (neighbor/chunk selections and weights), nor the on-device summaries are transmitted.

\subsection{Self-expressive Topology Reconstruction}
\label{sec:topo_recon}

\textbf{Motivation.}
After node/text condensation, many original edges are discarded, which will weaken message passing on the retained core.
To recover propagation capacity \emph{without} reverting to the full local graph, we reconstruct a lightweight adjacency on $\hat{V}^{(m)}_t$.

\textbf{Gated cross-modal fusion.}
For each $v\in\hat{V}^{(m)}_t$, we fuse the graph embedding $g_v$ and condensed evidence embedding $\tilde{t}_v$ into a unified feature $x_v\in\mathbb{R}^d$:

\begin{equation}
\label{eq:fused_embed}
\begin{aligned}
\alpha_v &= \sigma\!\big(w^\top[g_v\Vert \tilde{t}_v]\big),\\
x_v &= \mathrm{LN}\!\Big(W_g g_v + \alpha_v \cdot W_t \tilde{t}_v\Big),
\end{aligned}
\end{equation}
where $\alpha_v$ adaptively controls text injection and $\mathrm{LN}(\cdot)$ is layer normalization.Stacking $\{x_v\}_{v\in\hat V_t^{(m)}}$ yields the fused feature matrix $X$. We learn $\{W_g,W_t,w\}$ with supervised loss on the fused view and a lightweight cross-view consistency regularizer:
\begin{equation}
\label{eq:fusion_train_loss}
\begin{aligned}
\mathcal{L}_m
&=
\frac{1}{|\hat{V}^{(m)}_{t,L}|}
\sum_{v\in\hat{V}^{(m)}_{t,L}}
\mathrm{CE}\!\left(\mathrm{softmax}(o_v),y_v\right)\\
&+
\lambda_{\mathrm{align}}\,
\frac{1}{|\hat{V}^{(m)}_t|}
\sum_{v\in\hat{V}^{(m)}_t}
D\!\left(\mathrm{softmax}(o_v^{g}),\mathrm{softmax}(o_v^{t})\right)
\end{aligned}
\end{equation}
where $o_v=\mathrm{Dec}(x_v)$, $o_v^{g}=\mathrm{Dec}_g(g_v)$, and $o_v^{t}=\mathrm{Dec}_t(\tilde{t}_v)$ are logits from lightweight heads and $D(\cdot,\cdot)$ is a divergence (e.g., $\mathrm{KL}$).
The text encoder used to form $\tilde{t}_v$ remains frozen.

\textbf{Prior-regularized sparse self-expression.}
We reconstruct edges by sparse self-expression on fused features, while restricting coefficients to an evidence-aware candidate support.
Let $S_{ij}\in[0,1]$ be an evidence-induced prior (derived from chunk-level matching; cf.\ Sec.~\ref{sec:text_condense}), where larger $S_{ij}$ indicates stronger textual support between $i$ and $j$.
For each $i\in\hat{V}^{(m)}_t$, we construct a candidate set by combining fused-feature similarity and the evidence prior:
\begin{equation}
\label{eq:candidate_set}
\mathcal{C}(i) \triangleq \mathrm{TopK}_q\!\left(\{x_i^\top x_j\}_{j\neq i}\right)\ \cup\
\mathrm{TopK}_q\!\left(\{S_{ij}\}_{j\neq i}\right).
\end{equation}
We then optimize a sparse coefficient matrix $Z\in\mathbb{R}^{K\times K}$ (with $\mathrm{diag}(Z)=0$) supported on $\mathcal{C}(i)$:
\begin{equation}
\label{eq:self_expressive_sparse}
\begin{gathered}
\min_{Z}\;
\alpha\|X-XZ\|_F^2
+\beta\|Z\|_1
+\sum_{i\neq j}(1-S_{ij})|Z_{ij}|,\\
\text{s.t. }Z_{ij}=0\ (j\notin\mathcal{C}(i)).
\end{gathered}
\end{equation}
The prior term downweights unsupported links (small $S_{ij}$) by penalizing $|Z_{ij}|$ more strongly.
Restricting $Z$ to $\mathcal{C}(i)$ reduces dense $O(K^2)$ interactions to $O(Kq)$ nonzeros; we solve Eq.~\eqref{eq:self_expressive_sparse} with proximal gradient updates on the masked coefficients.

\textbf{Sparsified adjacency synthesis.}
We symmetrize and sparsify the reconstructed coefficients to obtain the propagation graph:
\begin{equation}
\label{eq:adjacency}
\begin{gathered}
W = |Z|+|Z|^\top,\\
\hat{A}^{(m)}_t = \mathcal{T}_k(W),
\end{gathered}
\end{equation}
where $\mathcal{T}_k(\cdot)$ keeps the top-$k$ entries per row (then symmetrizes), yielding an $O(Kk)$-edge graph for efficient message passing on the condensed core.

\section{Theoretical Analysis}
\label{sec:theory}
In this section, we provide a theoretical analysis of \textbf{DANCE} to justify its budgeted evidence design.
By enforcing \emph{hard} hop-wise and chunk-wise budgets, DANCE admits (i) deterministic sparsity and a clean cost decomposition,
(ii) bounded distortion introduced by hard projection, and (iii) stable evidence refresh under small model drift.
All proofs are deferred to Appendix~\ref{sec:proof}.
\subsection{Budget Guarantees and Cost Decomposition}
\label{sec:budget_cost_main}

\begin{definition}[End-to-end evidence processing under hard budgets]
\label{def:e2e_cost}
For client $m$, let $\hat{V}^{(m)}$ be the condensed core set with $|\hat{V}^{(m)}|=K$.
For each core node $v\in\hat{V}^{(m)}$ and hop $\ell\in\{0,1,2\}$, DANCE selects a neighbor set
$\mathcal{S}_v^{(\ell)}$ under budget $B_\ell$.
Let $\mathcal{E}_v$ be the candidate chunk set collected from the selected neighbors, and enforce a chunk budget
$\|\pi_{v,\cdot}\|_0\le B_{\mathrm{tok}}$.

We decompose the per-round cost on client $m$ as
\begin{equation}
\mathrm{Cost}^{(m)}=
\mathrm{Cost}_{\mathrm{agg}}^{(m)}+
\mathrm{Cost}_{\mathrm{backbone}}^{(m)}+
\mathrm{Cost}_{\mathrm{sel}}^{(m)},
\end{equation}
corresponding to backbone encoding, scoring/selection, and budgeted aggregation.
\end{definition}

\begin{theorem}[End-to-end complexity under hard budgets]
\label{thm:e2e_cost}
For any client $m$ and any stage, the hard budgets ensure
$|\mathcal{S}_v^{(\ell)}|\le B_\ell$ for $\ell\in\{0,1,2\}$ and $\|\pi_{v,\cdot}\|_0\le B_{\mathrm{tok}}$ for all
$v\in\hat{V}^{(m)}$, hence $\sum_{v\in\hat{V}^{(m)}}\|\pi_{v,\cdot}\|_0\le K B_{\mathrm{tok}}$.
Let $d$ be the embedding dimension. The per-round budgeted aggregation cost is
\begin{equation}
\label{eq:agg_cost_main}
\mathrm{Cost}_{\mathrm{agg}}^{(m)}
=
O\!\left(
dK\Big(\sum_{\ell\in\{0,1,2\}} B_\ell + B_{\mathrm{tok}}\Big)
\right).
\end{equation}
Therefore, the aggregation term is directly controlled by the hard budgets $\{B_\ell\}_{\ell\in\{0,1,2\}}$ and $B_{\mathrm{tok}}$.
Moreover, the backbone and scoring/selection terms are also budget-controlled via hard budgets(formal bounds in Appendix~\ref{cor:cost_v3}). Consequently, the end-to-end per-round cost on client $m$ is controlled under
hard budgets $\{B_\ell\}$ and $B_{\mathrm{tok}}$.

\end{theorem}

\begin{table*}[h]
    \centering
    \caption{Node classification accuracy (\%) comparison across datasets with varying condensation ratios.For methods with graph or data condensation, results are reported under the specified condensation ratios,
while methods without condensation are evaluated on the full datasets. The highest results are highlighted in \colorbox[HTML]{DADADA}{\textbf{bold}}, while the second-highest in \underline{underline}.}
    \label{tab:main_result}
    \vspace{-1pt}
    \resizebox{\textwidth}{!}{
    \begin{tabular}{c|c|c|c c c| c c c| c c c c c |c}
    \specialrule{1.5pt}{1.5pt}{1.5pt}
    
    \multicolumn{3}{c|}{\textbf{Setting}} &
    \multicolumn{12}{c}{\textbf{Methods}} \\
    \cmidrule{1-15}
    
    \textbf{Property} & \textbf{Dataset} & \textbf{Ratio} &
    \textbf{FedAvg}   & \textbf{FedSage+}  & \textbf{FedGTA}
    &\textbf{LLM4FGL} & \textbf{LLM4RGNN} & \textbf{LLaTA}
    & \textbf{GCond} & \textbf{SFGC} & \textbf{FedC4} &
    \textbf{FedGVD} &\textbf{FedGM}  & \textbf{Ours} \\
    \midrule
        \multirow{6}{*}{\textbf{Small}}
            &\cellcolor[HTML]{FFF0D5}\multirow{3}{*}
            & 8\%
            & 
            & 
            & 
            & 
            & 
            &
            & $83.61_{\scriptstyle \pm 0.26}$
            & $83.58_{\scriptstyle \pm 0.25}$
            & $86.02_{\scriptstyle \pm 0.21}$
            & $85.56_{\scriptstyle \pm 0.20}$
            & $84.15_{\scriptstyle \pm 0.13}$ 
            & \cellcolor[HTML]{DADADA}$\textbf{88.87}_{\scriptstyle \pm \textbf{0.43}}$ \\

            &\cellcolor[HTML]{FFF0D5}{\centering\textbf{Cora}}  & 4\%
            & $82.45_{\scriptstyle \pm 0.35}$
            & $83.37_{\scriptstyle \pm 0.54}$
            & $79.45_{\scriptstyle \pm 0.23}$
            & $86.73_{\scriptstyle \pm 0.42}$
            & \underline{$87.26_{\scriptstyle \pm 0.32}$}
            & $84.65_{\scriptstyle \pm 0.40}$
            & $82.97_{\scriptstyle \pm 0.40}$
            & $83.15_{\scriptstyle \pm 0.53}$
            & $84.98_{\scriptstyle \pm 0.42}$
            & $83.72_{\scriptstyle \pm 0.46}$
            & $80.78_{\scriptstyle \pm 0.41}$
            & {$87.12_{\scriptstyle \pm 0.42}$} \\
            &\cellcolor[HTML]{FFF0D5}  & 2\%
            & 
            & 
            & 
            & 
            & 
            & 
            & $81.36_{\scriptstyle \pm 0.61}$ 
            & $81.42_{\scriptstyle \pm 0.65}$
            & $84.33_{\scriptstyle \pm 0.56}$
            & $83.24_{\scriptstyle \pm 0.64}$
            & $74.09_{\scriptstyle \pm 0.54}$
            
            & $86.32_{\scriptstyle \pm 0.66}$ \\
            \cmidrule{2-15}
            & \cellcolor[HTML]{FFF7CD} \multirow{3}{*} 
            & 8\%
            & 
            & 
            & 
            & 
            & 
            & 
            & $72.03_{\scriptstyle \pm 0.27}$
            & $73.14_{\scriptstyle \pm 0.24}$
            & $75.95_{\scriptstyle \pm 0.22}$
            & $72.61_{\scriptstyle \pm 0.27}$
            & $73.82_{\scriptstyle \pm 0.21}$
            
            & \cellcolor[HTML]{DADADA}$\textbf{80.19}_{\scriptstyle \pm \textbf{0.76}}$ \\
            & \cellcolor[HTML]{FFF7CD}{\textbf{CiteSeer}}  
            & 4\%
            & $75.28_{\scriptstyle \pm 0.41}$
            & $75.12_{\scriptstyle \pm 0.51}$
            & $75.44_{\scriptstyle \pm 0.51}$
            & $75.53_{\scriptstyle \pm 0.22}$
            & $76.96_{\scriptstyle \pm 0.46}$
            & \underline{$78.21_{\scriptstyle \pm 0.43}$}
            & $67.46_{\scriptstyle \pm 0.40}$
            & $71.03_{\scriptstyle \pm 0.45}$
            & $74.84_{\scriptstyle \pm 0.36}$
            & $72.11_{\scriptstyle \pm 0.51}$
            & $72.57_{\scriptstyle \pm 0.43}$
            
            & {$78.19_{\scriptstyle \pm 0.48}$} \\
            & \cellcolor[HTML]{FFF7CD}
            & 2\%
            & 
            & 
            & 
            & 
            & 
            & 
            & $67.27_{\scriptstyle \pm 0.66}$
            & $69.34_{\scriptstyle \pm 0.65}$
            & $73.91_{\scriptstyle \pm 0.62}$
            & $67.22_{\scriptstyle \pm 0.57}$
            & $70.21_{\scriptstyle \pm 0.43}$
            
            & $75.93_{\scriptstyle \pm 0.63}$ \\
    \midrule
        \multirow{6}{*}{\textbf{Medium}}
            & \cellcolor[HTML]{FFF1B8} \multirow{3}{*}
            & 8\%
            &
            &
            & 
            & 
            &
            &
            & $68.92_{\scriptstyle \pm 0.31}$
            & $70.48_{\scriptstyle \pm 0.32}$
            & $72.87_{\scriptstyle \pm 0.24}$
            & $70.08_{\scriptstyle \pm 0.21}$
            & $71.72_{\scriptstyle \pm 0.23}$
            
            & \cellcolor[HTML]{DADADA}$\textbf{78.26}_{\scriptstyle \pm \textbf{0.26}}$ \\
            & \cellcolor[HTML]{FFF1B8}{\textbf{Arxiv}}
            & 4\%
            & $72.19_{\scriptstyle \pm 0.48}$
            & OOT
            & $71.18_{\scriptstyle \pm 0.41}$
            & $69.59_{\scriptstyle \pm 0.14}$
            & $70.72_{\scriptstyle \pm 0.35}$
            & $73.24_{\scriptstyle \pm 0.44}$
            & $67.23_{\scriptstyle \pm 0.41}$
            & $69.24_{\scriptstyle \pm 0.51}$
            & $71.44_{\scriptstyle \pm 0.36}$
            & $69.07_{\scriptstyle \pm 0.44}$
            & $69.83_{\scriptstyle \pm 0.56}$
            
            & \underline{$75.15_{\scriptstyle \pm 0.46}$} \\
            & \cellcolor[HTML]{FFF1B8}
            & 2\%
            &
            &
            &
            & 
            &
            &
            & $65.47_{\scriptstyle \pm 0.57}$
            & $67.24_{\scriptstyle \pm 0.61}$
            & $69.83_{\scriptstyle \pm 0.57}$
            & $67.22_{\scriptstyle \pm 0.60}$
            & $68.34_{\scriptstyle \pm 0.59}$
            
            & $73.36_{\scriptstyle \pm 0.56}$ \\
            \cmidrule{2-15}
            & \cellcolor[HTML]{F4FFB8} \multirow{3}{*}
            & 8\%
            &
            &
            &
            & 
            & 
            &
            & $79.33_{\scriptstyle \pm 0.25}$
            & $80.08_{\scriptstyle \pm 0.33}$
            & $81.27_{\scriptstyle \pm 0.21}$
            & $79.14_{\scriptstyle \pm 0.31}$
            & $79.92_{\scriptstyle \pm 0.30}$
             
            & \cellcolor[HTML]{DADADA}$\textbf{84.26}_{\scriptstyle \pm \textbf{0.35}}$ \\
            & \cellcolor[HTML]{F4FFB8}{\textbf{WikiCS}}  
            & 4\%
            & $76.26_{\scriptstyle \pm 0.43}$
            & $81.25_{\scriptstyle \pm 0.39}$
            & $79.35_{\scriptstyle \pm 0.54}$
            & $79.31_{\scriptstyle \pm 0.18}$
            & OOT
            & \underline{$81.58_{\scriptstyle \pm 0.39}$}
            & $78.04_{\scriptstyle \pm 0.53}$
            & $78.97_{\scriptstyle \pm 0.48}$
            & $80.63_{\scriptstyle \pm 0.43}$
            & $78.71_{\scriptstyle \pm 0.42}$
            & $79.02_{\scriptstyle \pm 0.52}$
            
            & $81.27_{\scriptstyle \pm 0.42}$ \\
            & \cellcolor[HTML]{F4FFB8}
            & 2\%
            &
            &
            &
            & 
            &
            &
            & $76.47_{\scriptstyle \pm 0.62}$
            & $77.34_{\scriptstyle \pm 0.68}$
            & $79.36_{\scriptstyle \pm 0.66}$
            & $76.99_{\scriptstyle \pm 0.57}$
            & $78.43_{\scriptstyle \pm 0.49}$
            
            & $80.58_{\scriptstyle \pm 0.56}$ \\
    \midrule
        \multirow{6}{*}{\textbf{Inductive}}
            & \cellcolor[HTML]{E6FDD1} \multirow{3}{*}
            & 2\%
            &
            &
            &
            & 
            & 
            &
            & $64.82_{\scriptstyle \pm 0.23}$
            & $65.12_{\scriptstyle \pm 0.34}$
            & $64.83_{\scriptstyle \pm 0.28}$
            & $63.52_{\scriptstyle \pm 0.26}$
            & $63.57_{\scriptstyle \pm 0.21}$
             
            & \cellcolor[HTML]{DADADA}$\textbf{65.17}_{\scriptstyle \pm \textbf{0.22}}$ \\
            & \cellcolor[HTML]{E6FDD1}{\textbf{Instagram}}  
            & 1\%
            & $64.15_{\scriptstyle \pm 0.51}$
            & $63.54_{\scriptstyle \pm 0.49}$
            & $64.07_{\scriptstyle \pm 0.53}$
            & $60.98_{\scriptstyle \pm 0.32}$
            & OOT
            & $64.53_{\scriptstyle \pm 0.55}$
            & $64.38_{\scriptstyle \pm 0.53}$
            & $64.12_{\scriptstyle \pm 0.44}$
            & $64.03_{\scriptstyle \pm 0.42}$
            & $62.64_{\scriptstyle \pm 0.44}$
            & $64.25_{\scriptstyle \pm 0.40}$
            
            & \underline{$64.84_{\scriptstyle \pm 0.58}$} \\
            & \cellcolor[HTML]{E6FDD1}
            & 0.5\%
            &
            &
            &
            & 
            &
            &
            & $65.84_{\scriptstyle \pm 0.58}$
            & $63.14_{\scriptstyle \pm 0.63}$
            & $63.55_{\scriptstyle \pm 0.58}$
            & $62.24_{\scriptstyle \pm 0.60}$
            & $63.09_{\scriptstyle \pm 0.46}$
            
            & $64.75_{\scriptstyle \pm 0.56}$ \\
            \cmidrule{2-15}
            & \cellcolor[HTML]{D9F7BE} \multirow{3}{*}
            & 8\%
            &
            &
            &
            & 
            & 
            &
            & $67.23_{\scriptstyle \pm 0.27}$
            & \underline{$67.96_{\scriptstyle \pm 0.20}$}
            & $66.08_{\scriptstyle \pm 0.31}$
            & $64.66_{\scriptstyle \pm 0.20}$
            & $65.73_{\scriptstyle \pm 0.45}$
             
            & \cellcolor[HTML]{DADADA}$\textbf{68.45}_{\scriptstyle \pm \textbf{0.32}}$ \\
            & \cellcolor[HTML]{D9F7BE}{\textbf{Reddit}}  & 4\%
            & $59.22_{\scriptstyle \pm 0.45}$
            & $63.28_{\scriptstyle \pm 0.39}$
            & $65.54_{\scriptstyle \pm 0.48}$
            & $64.32_{\scriptstyle \pm 0.45}$
            & OOT
            & $67.62_{\scriptstyle \pm 0.54}$
            & $65.76_{\scriptstyle \pm 0.47}$
            & $66.53_{\scriptstyle \pm 0.50}$
            & $65.24_{\scriptstyle \pm 0.42}$
            & $63.65_{\scriptstyle \pm 0.50}$
            & $64.45_{\scriptstyle \pm 0.49}$
            
            & {$67.88_{\scriptstyle \pm 0.41}$} \\
            & \cellcolor[HTML]{D9F7BE}
            & 2\%
            &
            &
            &
            & 
            &
            &
            & $63.94_{\scriptstyle \pm 0.57}$
            & $65.48_{\scriptstyle \pm 0.62}$
            & $64.35_{\scriptstyle \pm 0.64}$
            & $63.26_{\scriptstyle \pm 0.56}$
            & $63.54_{\scriptstyle \pm 0.51}$
            
            & $65.96_{\scriptstyle \pm 0.62}$ \\
        \midrule
        \multirow{3}{*}{\textbf{Large}}
            & \cellcolor[HTML]{E6F7FF} \multirow{3}{*}
            & 1\%
            &
            &
            &
            & 
            & 
            & 
            & $47.32_{\scriptstyle \pm 0.35}$
            & $46.83_{\scriptstyle \pm 0.21}$
            & \underline{$48.24_{\scriptstyle \pm 0.35}$}
            & OOT
            & $47.36_{\scriptstyle \pm 0.40}$
            
            & \cellcolor[HTML]{DADADA}$\textbf{48.25}_{\scriptstyle \pm \textbf{0.28}}$ \\
            & \cellcolor[HTML]{E6F7FF}{\textbf{Children}}  
            & 0.6\%
            & $42.02_{\scriptstyle \pm 0.27}$
            & $47.38_{\scriptstyle \pm 0.30}$
            & $47.16_{\scriptstyle \pm 0.36}$
            & OOT
            & OOT
            & $47.26_{\scriptstyle \pm 0.38}$
            & $44.87_{\scriptstyle \pm 0.27}$
            & $45.14_{\scriptstyle \pm 0.31}$
            & $47.88_{\scriptstyle \pm 0.37}$
            & OOT
            & $45.02_{\scriptstyle \pm 0.28}$
            
            & $47.17_{\scriptstyle \pm 0.35}$ \\
            & \cellcolor[HTML]{E6F7FF}
            & 0.4\%
            &
            &
            &
            & 
            &
            &
            & $42.54_{\scriptstyle \pm 0.43}$
            & $43.13_{\scriptstyle \pm 0.54}$
            & $47.53_{\scriptstyle \pm 0.52}$
            & OOT
            & $43.42_{\scriptstyle \pm 0.30}$
            
            & $46.18_{\scriptstyle \pm 0.46}$ \\
        \midrule
        \multirow{3}{*}{\textbf{Hetero.}}
            & \cellcolor[HTML]{F0F5FF} \multirow{3}{*}
            & 8\%
            &
            &
            &
            & 
            & 
            &
            & $43.03_{\scriptstyle \pm 0.63}$
            & $44.02_{\scriptstyle \pm 0.57}$
            & $45.24_{\scriptstyle \pm 0.70}$
            & $44.93_{\scriptstyle \pm 0.66}$
            & $44.63_{\scriptstyle \pm 0.62}$
             
            & \cellcolor[HTML]{DADADA}$\textbf{47.84}_{\scriptstyle \pm \textbf{0.61}}$ \\
            & \cellcolor[HTML]{F0F5FF}{\textbf{Ratings}}  
            & 6\%
            & $40.75_{\scriptstyle \pm 0.62}$
            & $45.98_{\scriptstyle \pm 0.66}$
            & $43.56_{\scriptstyle \pm 0.61}$
            & $42.23_{\scriptstyle \pm 0.52}$
            & OOT
            & $44.47_{\scriptstyle \pm 0.66}$
            & $42.12_{\scriptstyle \pm 0.70}$
            & $42.97_{\scriptstyle \pm 0.60}$
            & $44.13_{\scriptstyle \pm 0.56}$
            & $43.85_{\scriptstyle \pm 0.58}$
            & $44.02_{\scriptstyle \pm 0.61}$
            
            & \underline{$46.88_{\scriptstyle \pm 0.70}$} \\
            & \cellcolor[HTML]{F0F5FF}
            & 4\%
            &
            &
            & 
            &
            &
            &
            & $41.23_{\scriptstyle \pm 0.63}$
            & $42.14_{\scriptstyle \pm 0.57}$
            & $43.18_{\scriptstyle \pm 0.71}$
            & $43.41_{\scriptstyle \pm 0.66}$
            & $43.07_{\scriptstyle \pm 0.68}$
            
            & $45.43_{\scriptstyle \pm 0.61}$ \\   
    \specialrule{1.3pt}{2.0pt}{1.0pt}
    \end{tabular}}
    \vspace{-3pt}

    \label{table:node_cls}
    \end{table*}

\subsection{Approximation Error from Hard Budget Projection}
\label{sec:approx_err_main}

\begin{definition}[Tail mass under top-$B$ truncation]
\label{def:tail_mass_topB}
Let $p_v\in\Delta^{|\mathcal{E}_v|}$ be the \emph{untruncated} chunk-weight distribution produced by the same scoring rule (e.g., entmax),
and let $\pi_v=\Pi_{B_{\mathrm{tok}}}(p_v)$ be the hard-budgeted weights.
Define the tail mass beyond the top-$B_{\mathrm{tok}}$ entries as
\begin{equation}
\delta_{B_{\mathrm{tok}}}(p_v)\;\triangleq\;\sum_{i\notin T_{B_{\mathrm{tok}}}(p_v)} p_{v,i}.
\label{eq:tail_mass_main}
\end{equation}
Hard budgets restrict evidence size, but they need not arbitrarily distort the evidence representation.
Our analysis isolates the effect of \emph{hard projection} $\Pi_{B_{\mathrm{tok}}}$ by comparing against the same scoring rule without truncation.
\end{definition}

\begin{theorem}[Bounded distortion from hard truncation]
\label{thm:bounded_distortion_trunc}
Assume chunk embeddings are bounded as in Assumption~\ref{ass:bounded_rep}, i.e., $\|e_{u,r}\|_2\le M$.
Let
$t_v^{\mathrm{full}}=\sum_{(u,r)\in\mathcal{E}_v} (p_v)_{(u,r)}e_{u,r}$
and
$\tilde{t}_v=\sum_{(u,r)\in\mathcal{E}_v} (\pi_v)_{(u,r)}e_{u,r}$.
Then
\begin{equation}
\|\tilde{t}_v-t_v^{\mathrm{full}}\|_2 \;\le\; 2M\cdot \delta_{B_{\mathrm{tok}}}(p_v).
\label{eq:approx_bound_main}
\end{equation}
This is Proposition~\ref{prop:chunk_err_v3} (proof in Appendix~\ref{sec:approx_err}).
When the scoring distribution is already sparse (small tail mass), the additional hard truncation introduces only limited approximation error while enforcing strict budgets.
\end{theorem}

\subsection{Stability of Model-in-the-loop Evidence Refresh}
\label{sec:stability_main}

\begin{definition}[Top-$B$ margin for score stability]
\label{def:topB_margin}
Let $r(\omega)\in\mathbb{R}^n$ denote the \emph{pre-entmax} score vector (neighbor scores or chunk scores) under parameters $\omega$,
sorted as $r_{(1)}(\omega)\ge\cdots\ge r_{(n)}(\omega)$.
The top-$B$ margin is defined as
\begin{equation}
\Delta_B(\omega)\;\triangleq\; r_{(B)}(\omega)-r_{(B+1)}(\omega).
\label{eq:margin_main}
\end{equation}

Although DANCE refreshes scores using the evolving model, the selected evidence sets do not necessarily change every time:
selection changes only when the ordering near the budget boundary is perturbed enough to swap items across rank $B$.
\end{definition}

\begin{theorem}[Selection stability under bounded model drift]
\label{thm:selection_stability_drift}
Assume the score functions are locally Lipschitz along the trajectory as in Assumption~\ref{ass:lipschitz_score_v3}:
$|r_j(\omega)-r_j(\omega')|\le L_s\|\omega-\omega'\|_2$ for all coordinates $j$.
If
\begin{equation}
\|\omega-\omega'\|_2 \;\le\; \frac{\Delta_B(\omega)}{2L_s},
\label{eq:stability_cond_main}
\end{equation}
then the selected top-$B$ index set after entmax and truncation remains invariant
(Theorem~\ref{thm:stability_v3}, proof in Appendix~\ref{sec:stability}).
This formalizes \emph{controlled adaptivity}: evidence refresh reacts to meaningful score reordering, while remaining stable under small parameter drift.
\end{theorem}

\section{Experiments}
\label{sec:exp}

In this section, we provide a comprehensive empirical evaluation of the proposed \textbf{DANCE} framework by addressing the following key research questions:

\textbf{Q1} – Does DANCE improve overall accuracy over strong FGL baselines across diverse benchmarks?  
\textbf{Q2} – How interpretable are the decision-making processes in DANCE, and how does it provide human-readable evidence for regulatory compliance?  
\textbf{Q3} – How robust is DANCE to hyper-parameter choices, Non-IID data, and the number of clients?  
\textbf{Q4} – What efficiency gains does DANCE achieve in terms of overall cost compared to baseline federated graph learning methods?
\textbf{Q5} – Can DANCE be combined with standard privacy mechanisms while maintaining competitive performance and privacy protection?  
\subsection{Experiments Setup}
\textbf{Baselines and Evaluation Metrics.}
Following prior studies on federated graph learning, we benchmark DANCE against representative methods from four categories.
(1) FL/FGL includes FedAvg~\cite{fedavg}, FedSage+~\cite{zhang2021subgraph}, and FedGTA~\cite{li2024fedgta}, which represent strong federated learning and structure-aware federated graph learning methods under subgraph-based FL settings.
(2) Graph condensation includes GCond~\cite{jin2021graph} and SFGC~\cite{zheng2023structure}
(3) Federated graph condensation includes FedC4~\cite{fedc4}, FedGVD~\cite{dai2025fedgvd}, and FedGM~\cite{fedgm}, which perform  condensation without preserving readable textual attributes.
(4) LLM-based TAG-FGL includes LLM4RGNN~\cite{zhang2025can}, LLaTA~\cite{zhang2025rethinking}, and LLM4FGL~\cite{yan2025llm4fgl}, which leverage large language models to enhance or augment text-attributed graphs.Detailed descriptions of each baseline and the rationale for their inclusion are provided in Appendix~\ref{sec:baseline_info}.

\textbf{Datasets and Federated Simulation.}
We evaluate DANCE on text-attributed graph (TAG) benchmarks spanning graph scales and multiple domains.
Additionally, we include heterophilic dataset to further validate DANCE under graph heterogeneity.
We follow the standard subgraph-FL protocol~\cite{zhang2021subgraph} to partition each global TAG into client subgraphs (e.g., Louvain).
Dataset statistics and default settings are reported in Table~\ref{tab:datasets}.

\subsection{Overall Performance (Q1)}
\label{sec:q1_overall}

\begin{figure}[t]
    \centering
    \captionsetup{skip=2pt}

    \includegraphics[width=0.49\columnwidth]{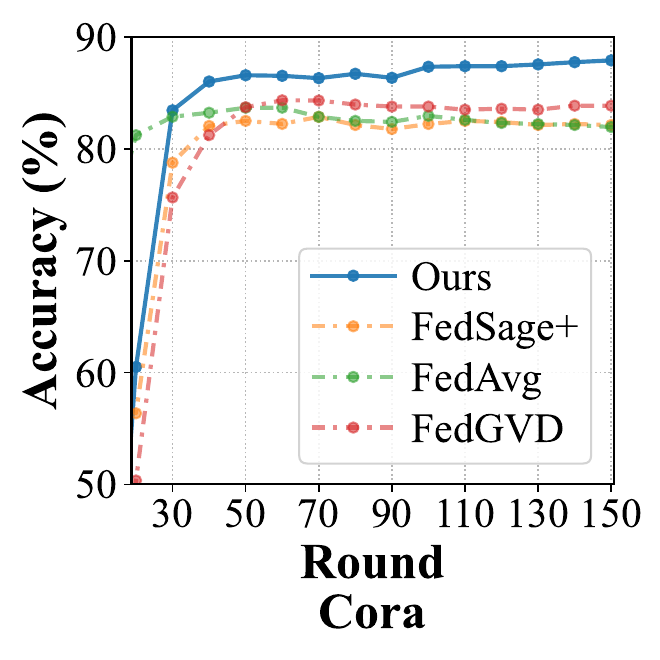}
    \hfill
    \includegraphics[width=0.49\columnwidth]{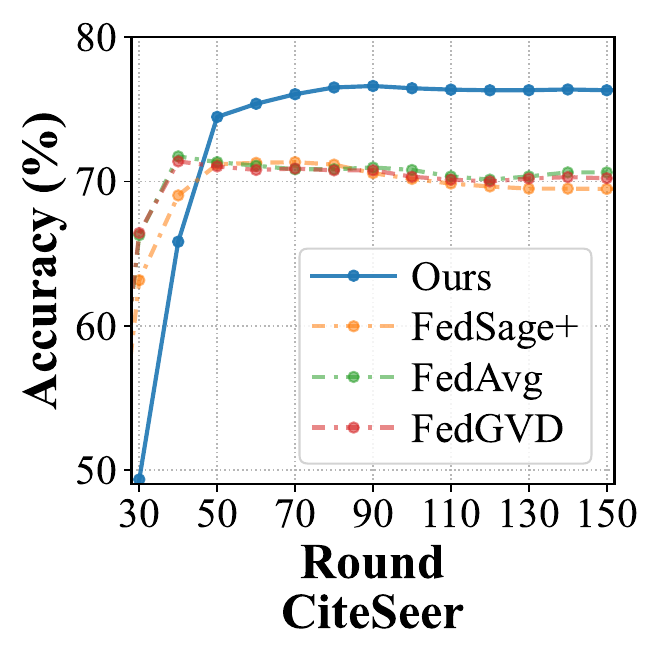}

    \caption{Convergence across communication rounds on Cora (left) and CiteSeer (right).}
    \label{fig:convergence}
    \vspace{-15pt}
\end{figure}

\textbf{Overall Results.}
To answer \textbf{Q1}, Table~\ref{tab:main_result} compares DANCE with all baselines across benchmarks.
DANCE consistently achieves the best or second-best Accuracy across benchmarks, even under compact condensation settings, suggesting that the proposed framework can improve predictive performance without merely trading accuracy for efficiency.We further report convergence in Figure~\ref{fig:convergence}, showing that DANCE reaches competitive performance earlier in training.Detailed experimental settings and implementation specifics are provided in
Appendix~\ref{sec:performance}.

\textbf{Ablation Study.}
To better understand the source of performance gains in DANCE,
we conduct an ablation study by removing each core component in turn.
As shown in Table~\ref{tab:ablation}, removing any core component leads to a consistent
performance degradation across datasets.
In particular, removing node condensation causes the largest drop,
highlighting the importance of jointly condensing structure and text
under federated non-IID settings.

\begin{table}[t]
    \centering
    \caption{Ablation study on Cora and CiteSeer datasets. Each variant removes one core module to validate its contribution. Bold values indicate the best performance in each column.}
    \label{tab:ablation}
    \vspace{-0pt}
    \resizebox{\columnwidth}{!}{
    \begin{tabular}{c|cc|cc}
    \specialrule{1.5pt}{1.5pt}{1.5pt}
    \multirow{2}{*}{\textbf{Datasets}} & 
    \multicolumn{2}{c|}{\textbf{Cora(0.08)}} & 
    \multicolumn{2}{c}{\textbf{CiteSeer(0.08)}} \\
    \cmidrule{2-5}
    & \textbf{Overall-F1} & \textbf{Acc}  & \textbf{Overall-F1} & \textbf{Acc} \\
    \midrule

     Full 
    & \cellcolor[HTML]{DADADA}$\textbf{88.36}_{\scriptstyle \pm \textbf{0.52}}$ 
    & \cellcolor[HTML]{DADADA}$\textbf{88.87}_{\scriptstyle \pm \textbf{0.43}}$ 

    & \cellcolor[HTML]{DADADA}$\textbf{77.16}_{\scriptstyle \pm \textbf{0.83}}$ 
    & \cellcolor[HTML]{DADADA}$\textbf{80.19}_{\scriptstyle \pm \textbf{0.76}}$  \\
    
     w/o TGR 
    & \underline{$86.32_{\scriptstyle \pm 0.64}$} 
    & \underline{$87.08_{\scriptstyle \pm 0.52}$} 
    & \underline{$73.32_{\scriptstyle \pm 0.88}$} 
    & \underline{$77.32_{\scriptstyle \pm 0.82}$} \\
    
     w/o TC
    & $82.92_{\scriptstyle \pm 0.71}$ 
    & $85.14_{\scriptstyle \pm 0.61}$ 
    & $73.07_{\scriptstyle \pm 1.02}$ 
    & $77.11_{\scriptstyle \pm 0.94}$ \\

     w/o NC 
    & $81.04_{\scriptstyle \pm 0.68}$ 
    & $81.22_{\scriptstyle \pm 0.56}$ 
    & $65.44_{\scriptstyle \pm 0.95}$ 
    & $75.19_{\scriptstyle \pm 0.89}$ \\

    \specialrule{1.3pt}{2.0pt}{1.0pt}
    \end{tabular}}
    \vspace{-3pt}
\end{table}

\subsection{Interpretability and Auditability Analysis (Q2)}
\label{sec:q2_interp}
To answer \textbf{Q2}, we conduct a qualitative case study and a complementary quantitative analysis to evaluate the interpretability of DANCE’s decision-making process and the human-readability of its extracted evidence, using faithfulness, sufficiency, and compactness as evaluation criteria.

\textbf{Qualitative Case Studies.}
We provide detailed qualitative case studies in Appendix~\ref{sec:case_cora}, illustrating how
DANCE produces concise, human-readable evidence packs and neighbor-level contribution traces.
This example demonstrate faithful evidence selection via paired sufficiency and comprehensiveness analyses, supporting compliance-oriented inspection while respecting
federated privacy constraints.

\textbf{Faithfulness, sufficiency, and compactness.}
We quantify faithfulness with both \emph{deletion} and \emph{insertion} tests: removing (or adding) the top-$k$ gated neighbor chunks and measuring the change in accuracy. We further measure \emph{sufficiency} by re-predicting using only the selected evidence chunks (keeping the graph backbone fixed) and comparing against using all available chunks, and \emph{compactness} by reporting the average numbers of selected neighbors/chunks/tokens. As controls, we compare against random selection under the same budgets and attention-weight baselines that treat dense attention scores as explanations.
Detailed experimental settings and implementation specifics are provided in Appendix~\ref{sec:case_cora}.
\subsection{Robustness Analysis (Q3)}
To answer \textbf{Q3}, we evaluate the robustness of DANCE under varying hyper-parameters, federation scales, and model refresh strategies, highlighting its robustness under different experimental conditions.
\label{sec:q3_robust}

\begin{figure}[t]
    \centering
    \includegraphics[width=\columnwidth]{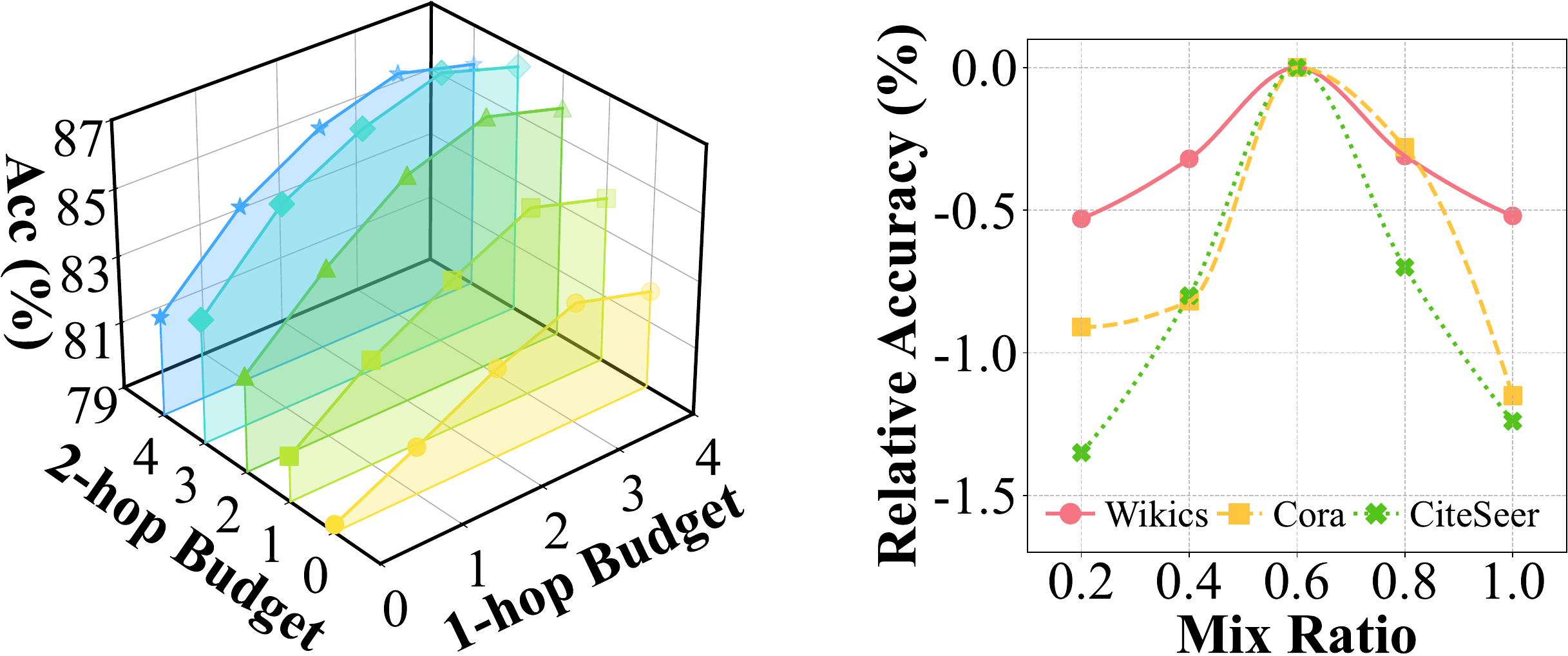}
    \vspace{-14pt}
    \caption{Hyper-parameter analysis of the \emph{Hierarchical Text Condensation} module.
    \textbf{Left:} accuracy under different 1-hop and 2-hop text budgets used in hierarchical evidence aggregation.
    \textbf{Right:} relative accuracy w.r.t.\ the summary mixing ratio $\mathit{mix}$, which balances condensed textual evidence and structural context.}
    \label{fig:hyper_text}
    \vspace{-4pt} 
\end{figure}

\begin{figure}[t]
    \centering
    \includegraphics[width=\columnwidth]{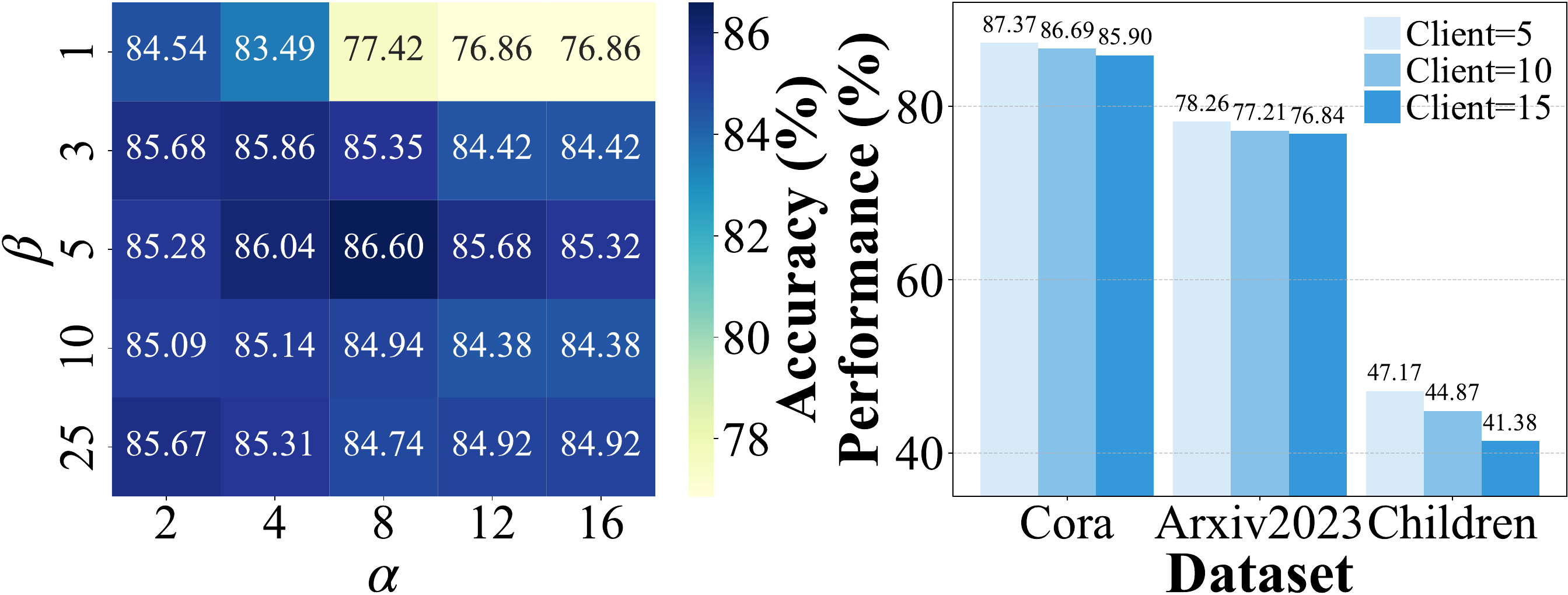}
    \vspace{-18pt}
    \caption{Hyper-parameter sensitivity and scalability analysis of the \emph{Self-expressive Topology Reconstruction} module.
    \textbf{Left:} node classification accuracy under different regularization weights $\alpha$ and $\beta$ in the prior-regularized sparse self-expression objective.
    \textbf{Right:} scalability with varying numbers of participating clients.}
    
    \vspace{-23pt}
    \label{fig:hyper_tgr_client}
\end{figure}

\textbf{Hyper-parameter Sensitivity.}
We examine the sensitivity of DANCE to the key hyper-parameters in \emph{Hierarchical Text Condensation} and \emph{Self-expressive Topology Reconstruction}.
As shown in Fig.~\ref{fig:hyper_text}, DANCE remains stable across a wide range of 1-hop/2-hop evidence budgets and the summary mixing ratio $\mathit{mix}$, with a clear high-performing region.
Fig.~\ref{fig:hyper_tgr_client} further shows a smooth accuracy landscape when varying the regularization weights $\alpha$ and $\beta$ in Eq.~\eqref{eq:self_expressive_sparse}, indicating that the topology reconstruction is not overly sensitive to precise regularization choices.
We further evaluate DANCE across different GNN backbones in Appendix~\ref{app:backbone_robust}.

\textbf{Scaling Clients.}
We vary the number of participating subgraph clients $M$ to evaluate the stability of DANCE under different federation scales.
Figure~\ref{fig:hyper_tgr_client}(right) reports the results, showing that DANCE maintains stable and competitive performance as the number of clients increases.

\textbf{Model-in-the-Loop Refresh Study.}
We additionally isolate the effect of \emph{round-adaptive refresh} by comparing DANCE against (i) \textsc{Static-DANCE}, which performs condensation/evidence selection only once at $t{=}1$ and then keeps them fixed, (ii) \textsc{Refresh-Core-Only}, and (iii) \textsc{Refresh-Text-Only}. This directly tests whether model-in-the-loop refresh is necessary under non-IID federation.Detailed experimental settings are provided in Appendix~\ref{app:refresh_study}.

\subsection{Efficiency Analysis (Q4)}
\label{sec:q4_eff}

To answer \textbf{Q4}, we conduct both a theoretical and empirical efficiency analysis of DANCE,
aiming to demonstrate its computational and runtime advantages in federated text-attributed graph learning.

\textbf{Theoretical Complexity Analysis.}
We analyze the theoretical \emph{time and space} complexity of DANCE and compare it with representative federated and LLM-assisted baselines, including FedAvg, FedC4, FedTAD, LLM4RGNN, and LLaTA.
Table~\ref{tab:complexity} summarizes the corresponding costs.
Let $C$ denote the number of clients, $n$ the number of local nodes, and $K$ the number of condensed nodes per client.

For DANCE, the \emph{time complexity} consists of three components.
\textbf{Label-aware Node Condensation} incurs $\mathcal{O}(n\log n)$,
\textbf{Hierarchical Text Condensation} operates on the condensed core with $\mathcal{O}(K)$ cost,
and \textbf{Self-expressive Topology Reconstruction} introduces $\mathcal{O}(Kq)$ complexity.
Aggregated over $C$ clients, the overall time complexity is $\mathcal{O}(C(n\log n + Kq))$.
The corresponding \emph{space complexity} is $\mathcal{O}(n + Kq)$ per client.

\begin{table}[t]
\centering
\caption{Theoretical time and space complexity comparison of different methods.}
\label{tab:complexity}
\resizebox{0.9\columnwidth}{!}{%
\begin{tabular}{l c c}
\specialrule{1.5pt}{1.5pt}{1.5pt}
\textbf{Method} 
& \textbf{Time Complexity} 
& \textbf{Space Complexity} \\
\midrule
FedAvg   
& $\mathcal{O}(C n)$ 
& $\mathcal{O}(n)$ \\

FedC4    
& $\mathcal{O}\!\left(C(n\log n + Kq)\right)$ 
& $\mathcal{O}(n + Kq)$ \\

FedTAD   
& $\mathcal{O}(C n + n^2)$ 
& $\mathcal{O}(n^2)$ \\

LLM4RGNN 
& $\mathcal{O}(C(n^2 + n))$ 
& $\mathcal{O}(n^2)$ \\

LLaTA    
& $\mathcal{O}\!\left(C n \log^2 n + (C^2 + 1)n\right)$ 
& $\mathcal{O}(n + Cn)$ \\

\midrule
\textbf{DANCE} 
& $\mathcal{O}\!\left(C(n\log n + Kq)\right)$ 
& $\mathcal{O}(n + Kq)$ \\
\specialrule{1.5pt}{1.5pt}{1.5pt}
\end{tabular}}
\end{table}

\begin{figure}[t]
    \centering
    \includegraphics[width=\columnwidth]{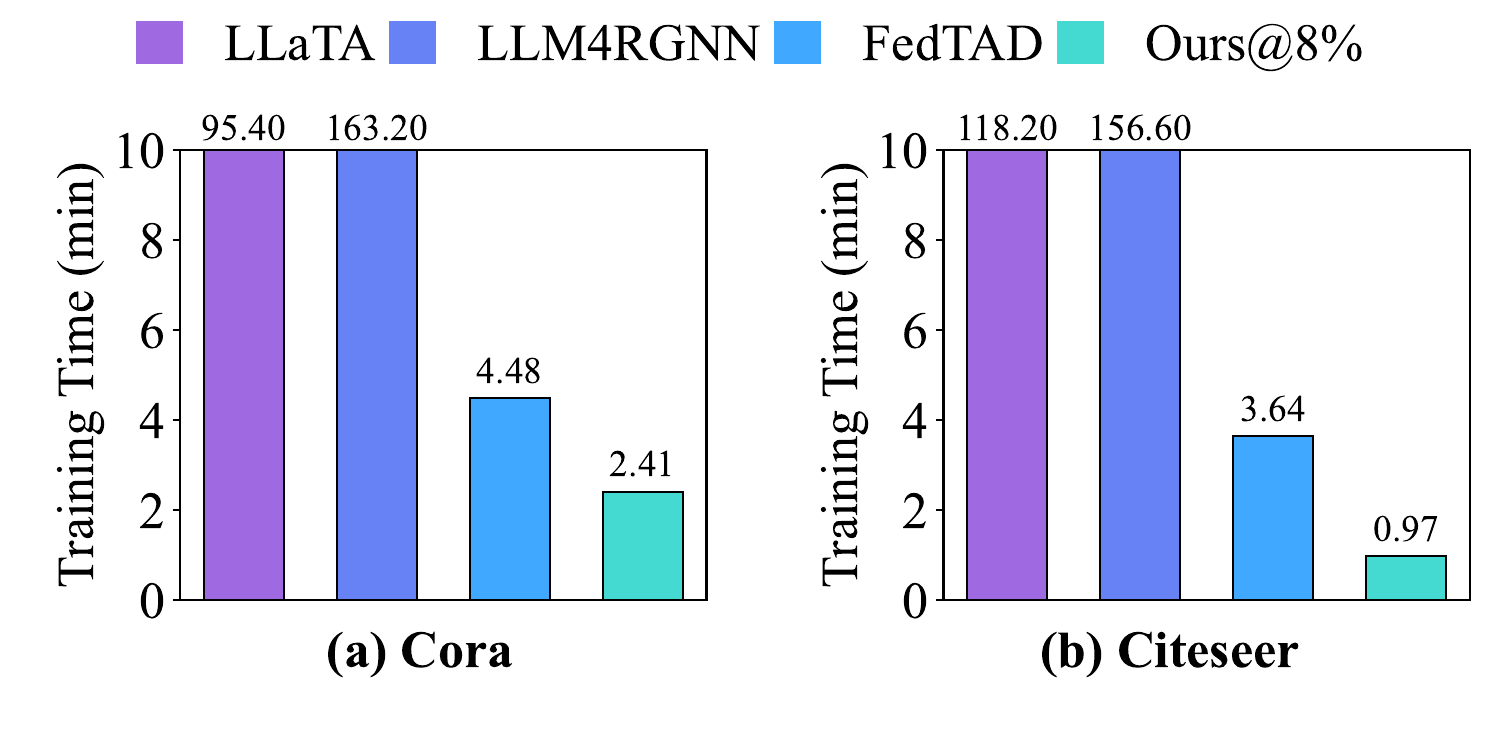}
    \vspace{-34pt}
    \caption{Training time comparison on Cora and Citeseer.
    DANCE achieves substantially lower runtime than LLM-based methods and also yields noticeable reductions compared to conventional federated baselines.
    }
    \vspace{-4pt}
    \label{fig:runtime}
\end{figure}

\textbf{Empirical Runtime Evaluation.}
We evaluate the efficiency of DANCE by comparing training time against representative baselines.Figure~\ref{fig:runtime} reports results on Cora and Citeseer.
LLM-based methods (e.g., LLaTA and LLM4RGNN) incur higher cost due to repeated LLM invocations, while FedTAD reduces overhead by offloading generation to the server but remains non-trivial.
In contrast, DANCE achieves significantly lower training time under the same federated setting.

\subsection{Privacy Analysis (Q5)}
\label{sec:q5_privacy}
To answer \textbf{Q5}, we apply secure aggregation~\cite{secagg} to uploaded client updates and any server-side coordination signals.
Since raw texts and decision traces are never transmitted, DANCE remains unchanged in how it generates local audit evidence.
\begin{figure}[H]
    \centering
    \includegraphics[width=0.6\columnwidth]{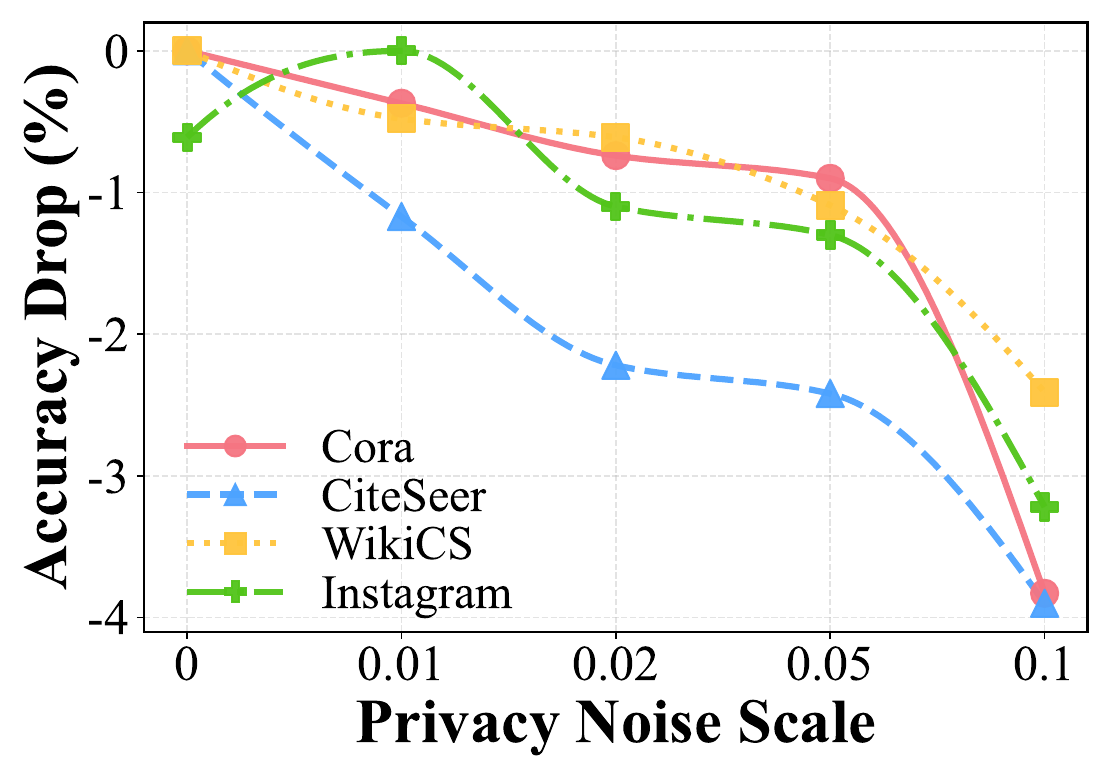}
    \vspace{-8pt}
\caption{Privacy--utility trade-off under different noise budgets.
DANCE retains competitive performance even with large noise}
    \vspace{-4pt}
    \label{fig:privacy}
\end{figure}
\textbf{Differential Privacy.}
We further evaluate DANCE under differential privacy by injecting Laplace noise into node representations.Figure~\ref{fig:privacy} reports the resulting privacy--utility trade-off across different noise scales.DANCE maintains competitive performance under moderate noise levels, suggesting that the proposed framework remains practically effective under standard privacy perturbations.
Implementation details are provided in Appendix~\ref{app:privacy_impl}.

\section{Conclusion}
We presented \textbf{DANCE}, a round-adaptive condensation framework for federated learning on text-attributed graphs.
By jointly budgeting node scale, neighbor usage, and chunk-level textual evidence, DANCE avoids repeatedly encoding and propagating redundant multi-hop texts, while refreshing retained evidence across rounds as the global model evolves under non-IID clients.
DANCE communicates only standard model updates, remaining compatible with secure aggregation and optional differential privacy, yet provides \emph{locally available} audit traces by materializing neighbor cues and supporting text chunks on-device.
Empirically, DANCE achieves consistently stronger node classification performance and calibration than competitive TAG-FGL baselines under substantially reduced token processing and communication budgets.
Future work includes more standardized interpretability and auditability evaluation, extending the framework to heterogeneous/temporal and multimodal graphs, and characterizing privacy--utility trade-offs under tighter accounting and stronger threat models.

\bibliography{example_paper}
\bibliographystyle{icml2025}

\newpage
\appendix
\onecolumn




\appendix
\onecolumn
\section{Methodology in Details}
\textbf{Difficulty scores for budgeted evidence.}
To prioritize informative (hard) evidence under tight budgets, we define a model-conditioned difficulty score for each node based on predictive uncertainty from the last-round global model:
\begin{equation}
\label{eq:difficulty}
u_v \triangleq \mathrm{H}\!\left(p^{(t-1)}(y\mid v)\right)
= -\sum_{c=1}^{C} p^{(t-1)}(c\mid v)\log p^{(t-1)}(c\mid v),
\end{equation}
where larger $u_v$ indicates higher uncertainty (harder nodes). We use $\{u_v\}$ only for evidence prioritization (e.g., selecting 2-hop candidates) and do not treat it as a pseudo-label signal.

\section{Experiment in Details}

\subsection{Computation Resource}
Experiments are conducted on a Linux server equipped with 2 Intel(R) Xeon(R) Gold 6240 CPUs @ 2.60GHz (36 cores per socket, 72 threads total), 251GB RAM (with approximately 216GB available), and 4 NVIDIA A100 GPUs with 40GB memory each. The software environment includes Python 3.10.19 and PyTorch 2.4.1 with CUDA 12.4. We adopt a multi-GPU training setup where each simulated client is assigned to a dedicated GPU during federated optimization.
\subsection{Overall Performance: Experimental Settings and Protocol}
\label{sec:performance}

This subsection details the experimental settings and evaluation protocol used for the overall
performance comparison in Section~\ref{sec:q1_overall}.

\textbf{Unified baseline adaptation.}
For graph condensation methods originally proposed in centralized settings, we apply condensation
\emph{locally} on each client’s subgraph under the federated setting.
The condensed representations are then used within the same federated training and aggregation
pipeline as other methods.
For baselines not specifically designed for text-attributed graphs, we adopt the same SBERT encoder
as DANCE to obtain node text representations, ensuring a consistent text processing backbone across
methods.

This unified treatment guarantees that observed performance differences stem from the learning,
condensation, and aggregation strategies, rather than discrepancies in textual feature extraction
or encoder capacity.

\textbf{Evaluation metrics.}
We report \textbf{Accuracy} as the primary predictive metric, following standard practice in
semi-supervised node classification on citation and text-rich graphs.
Additional metrics such as macro-F1 and heterogeneity-aware scores are reported where appropriate
to capture performance under non-IID federated settings.

\textbf{Federated training protocol.}
Unless otherwise stated, all methods are evaluated under the same federated schedule, including
identical client partitions, communication rounds, and local training epochs.
Hyper-parameters for all methods are selected on a validation split using a matched tuning budget
to avoid unfair optimization advantages.

\textbf{Randomness and reporting.}
All reported results are averaged over five independent random seeds and presented as
mean$\pm$standard deviation.
For convergence analysis, we track test accuracy across communication rounds using the same
evaluation checkpoints for all methods.

Overall, this protocol ensures a fair and controlled comparison across heterogeneous baselines,
isolating the impact of condensation, neighbor-gated aggregation, and evidence selection mechanisms
introduced by DANCE.

\subsection{Case Study: Faithful Evidence Selection on Cora}
\label{sec:case_cora}

\textbf{Setup.}
We present an illustrative qualitative example on the Cora citation network to examine evidence selection at the chunk level.No manual filtering or post hoc modification is applied.The same automated pipeline as in quantitative experiments is used, while summary-level sufficiency and comprehensiveness are evaluated separately in subsequent experiments.

\begin{figure}[t]
\vspace{-0.5em}
\centering
\small
\begin{tabular}{p{0.46\linewidth} p{0.46\linewidth}}
\toprule
\textbf{Original Paper (Excerpt)} &
\textbf{Selected Evidence Chunks by DANCE} \\
\midrule

\textbf{Title:} Exploration and Model Building in Mobile Robot Domains

\vspace{0.3em}
\textbf{Label: Neural Networks}

\vspace{0.6em}
\textbf{Abstract (excerpt):}

\emph{Real-world experiences are generalized via two artificial neural networks that encode the characteristics of the robot's sensors, as well as the characteristics of typical environments the robot is assumed to face.}

\emph{Once trained, these networks allow for knowledge transfer across different environments ...}

\emph{... an efficient dynamic programming method is employed in background ...}
&
\textbf{Evidence selected in core node:}
\begin{itemize}
    \item ``Real-world experiences are generalized via two artificial \underline{neural networks} that encode the characteristics of the robot's sensors.''
    \item ...
\end{itemize}

\vspace{0.2em}
\textbf{Evidence selected in hop neighbors:}
\begin{itemize}
    \item ``The resulting learning system is more stable and effective in changing environments than plain \underline{backpropagation}...''
    \item ...
\end{itemize}
\\
\bottomrule
\end{tabular}

\vspace{0.6em}
\caption{
\textbf{Illustrative example on the Cora dataset (Neural Networks class).}
The original abstract contains extensive descriptions related to robotics, neural networks, and system-level details.
DANCE selectively extracts evidence that is directly relevant to the node prediction. The selected evidence is presented as intact, human-readable text.
}
\label{fig:case_study}
\vspace{-1em}
\end{figure}

\textbf{Evidence selection.}
Evidence selection is performed automatically by DANCE through neighbor-gated aggregation and
chunk-level selection.
Each chunk is embedded and scored based on its contribution to the node representation under
neighbor-aware message aggregation.
The final evidence set is obtained by selecting the top-ranked chunks under the same budget
constraints used at test time.

In this example, DANCE selects human-readable chunks from the core node that highly relevant to neural network.
It further retrieves evidence chunks from hop neighbors that do not contain the term ``neural networks'', but are also highly relevant to neural networks.At the same time, other only loosely related content, such as mentions of dynamic programming methods, is not selected, reflecting DANCE’s preference for evidence that is more directly aligned with the prediction task.

\begin{figure}[H]
    \centering
    \includegraphics[width=0.44\linewidth]{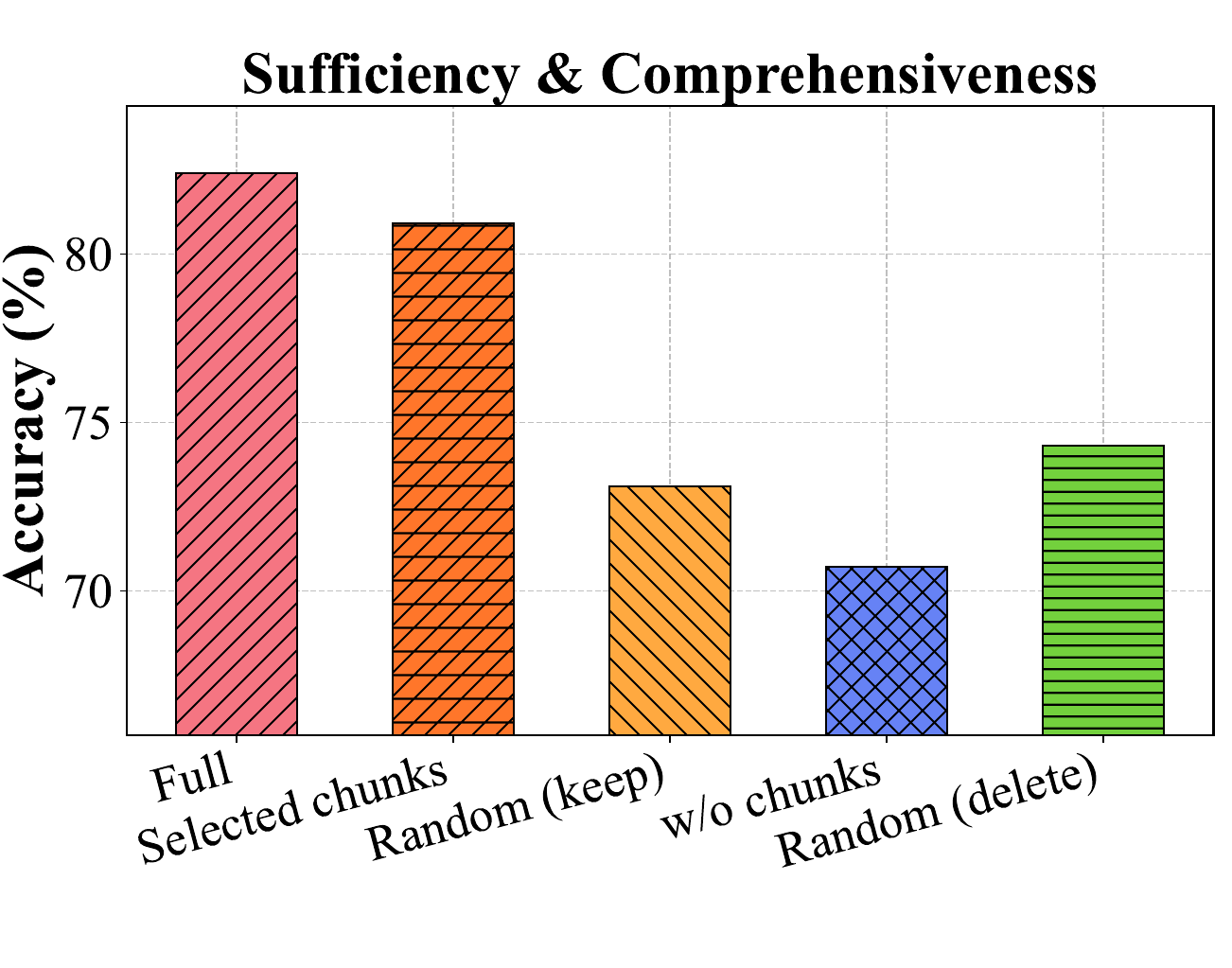}
    \vspace{-3pt}
    \caption{
    Chunk-level faithfulness analysis on Cora.
    The plot compares model accuracy under different evidence conditions, including the full text, retaining only DANCE-selected chunks (sufficiency), removing the selected chunks (comprehensiveness), and corresponding random controls with the same budget.
    }

    \label{fig:cora_faithfulness}
\end{figure}

\textbf{Faithfulness evaluation protocol.}
We evaluate faithfulness using paired \emph{sufficiency} and \emph{comprehensiveness} tests, following standard practice.
For sufficiency, we retain only the DANCE-selected chunks and mask all remaining text.
For comprehensiveness, we remove the selected evidence while keeping all other text intact.
Random chunk baselines of equal length are used as controls.
As shown in Figure~\ref{fig:cora_faithfulness}, retaining only the selected chunks largely preserves predictive performance, whereas removing them results in a noticeably larger degradation than random deletion, indicating that the selected evidence is both sufficient and necessary for the model’s prediction.

\textbf{Takeaway.}
This case study demonstrates that DANCE selects evidence that is both sufficient and necessary for
prediction under the same automated pipeline used in evaluation.
The selected evidence remains compact, human-interpretable, and faithful to the model’s actual
decision process, rather than reflecting post hoc or naively keyword-driven explanations.
\subsection{Robustness to Different GNN Backbones}
\label{app:backbone_robust}

\subsection{Model-in-the-Loop Refresh Study}
\label{app:refresh_study}
\paragraph{Experimental Setup.}
We study the effect of round-adaptive refresh under the standard subgraph federated learning setting.
All variants share the same federated schedule, client partition, communication rounds, and training hyper-parameters.
Unless otherwise stated, we follow the default configuration used in Section~\ref{sec:exp}, including the same condensation ratio, neighbor budgets, and chunk budgets.
\begin{figure}[t]
    \centering
    \includegraphics[width=0.65\columnwidth]{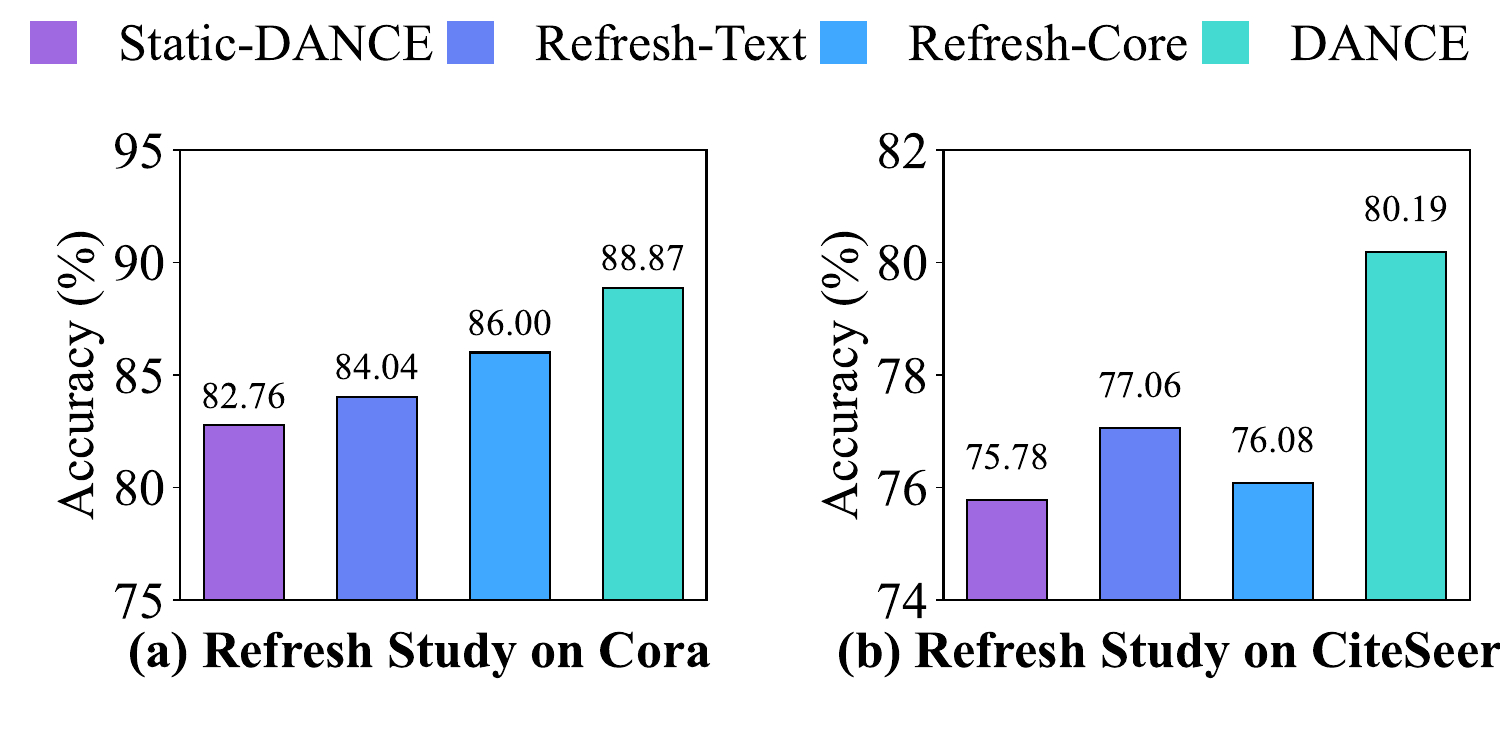}
    \caption{Model-in-the-loop refresh study on Cora and CiteSeer. 
    Full DANCE refreshes both the condensed node core and textual evidence at every round, while other variants refresh only partial components or remain static.}
    \label{fig:refresh_study}
\end{figure}

\paragraph{Compared Variants.}
We consider the following four methods:
\textbf{DANCE}: The full method, which refreshes both the condensed node core and evidence chunks at every round using the current global model.
\textbf{Static-DANCE}: Condensation and evidence selection are performed only once at $t{=}1$, and the resulting condensed graph and evidence are reused for all subsequent rounds.
\textbf{Refresh-Core-Only}: The condensed node core is refreshed at each round, while textual evidence (neighbor selection and chunk selection) is kept fixed after initialization.
\textbf{Refresh-Text-Only}: Textual evidence is refreshed at each round, while the condensed node core remains fixed.

\paragraph{Datasets and Metrics.}
We report node classification accuracy on Cora and CiteSeer, which are representative citation networks with non-IID client partitions induced by graph-based splitting.
Results are averaged over five random seeds and evaluated using the same test splits as in the main experiments.

\paragraph{Results and Analysis.}
Figure~\ref{fig:refresh_study} summarizes the results.
On both datasets, full DANCE outperforms Static-DANCE with an improvement of $+6.1\%$ on Cora and $+4.4\%$ on CiteSeer, demonstrating that condensation and evidence selection must be refreshed as the global model evolves.Partial-refresh variants recover part of the performance, but remain inferior to full DANCE.These results confirm that model-in-the-loop refresh of both structural and textual components is necessary under non-IID federated learning.

\subsection{Differential Privacy Implementation Details}
\label{app:privacy_impl}

We describe the concrete implementation of differential privacy used in the experiments reported in Section~\ref{sec:q5_privacy}.

\textbf{Noise injection mechanism.}
At each communication round, we perturb the uploaded condensed node representations on the client side before aggregation.
Specifically, we add element-wise Laplace noise to the condensed feature matrix:
\[
\tilde{\mathbf{X}} = \mathbf{X} + \mathrm{Laplace}(0, b),
\]
where the noise scale $b$ is determined by the target privacy budget~$\varepsilon$.

\textbf{Normalization and stability.}
After noise injection, the perturbed representations are $\ell_2$-normalized to prevent scale explosion and to stabilize subsequent aggregation and training.
No noise is added to raw texts, intermediate decision traces, or local evidence packs, which are never transmitted off-device.

\textbf{Scope of privacy protection.}
The applied differential privacy mechanism protects condensed representations exchanged during federated training.
Local audit artifacts, including neighbor importance scores and selected evidence chunks, remain entirely local and are unaffected by privacy perturbation.

\subsection{Datasets \& Simulation Method}
We evaluate \textbf{DANCE} on eight widely adopted text-attributed graph (TAG) datasets spanning multiple domains, including citation networks, one knowledge network, two social networks, and two e-commerce networks.
Among these datasets, the e-commerce datasets exhibit stronger heterophily, while the remaining datasets are relatively homophilic.
The details of these TAG datasets are summarized in Table~\ref{tab:datasets}.The description of the datasets for each domain is as follows:

\textbf{Citation Networks.}
    Cora, Citeseer, and Arxiv2023 are benchmark citation network datasets.
    Nodes represent academic papers, and edges denote citation relationships.
    Node texts consist of paper titles and abstracts, and labels correspond to research areas or subject categories.

\textbf{Knowledge Network.}
    WikiCS is a benchmark dataset of knowledge networks constructed from Wikipedia articles in computer science.
    Nodes represent articles, and edges indicate hyperlink relationships between articles.
    Node texts are obtained from article titles and abstracts, and labels denote different branches of computer science.

\textbf{Social Networks.}
    Instagram and Reddit are benchmark datasets of social networks.
    Nodes represent users, and edges represent social relationships.
    Textual features are derived from user profiles or recent posts, and labels correspond to different types of users or communities.

\textbf{E-commerce Networks.}
    Ratings and Children are benchmark datasets of e-commerce networks.
    Nodes represent products, and edges indicate relations such as co-purchase or co-view.
    Textual attributes include product names and descriptions, while labels correspond to product categories or rating groups.

\begin{table}[H]
\centering
\caption{Statistics of Datasets.}
\label{tab:datasets}
\begin{tabular}{lrrrrcll}
\specialrule{1.5pt}{1.5pt}{1.5pt}
\textbf{Dataset} 
& \textbf{Nodes} 
& \textbf{Edges} 
& \textbf{Classes} 
& \textbf{Train/Val/Test} 
& \textbf{Text Information} 
& \textbf{Domain} \\
\midrule
Cora        & 2,708  & 10,556 & 7  & 60/20/20    & Title and Abstract of Paper      & Citation \\
Citeseer    & 3,186  & 8,450  & 6  & 60/20/20    & Title and Abstract of Paper      & Citation \\
WikiCS      & 11,701 & 431,726 & 10 & 5/22.5/50  & Title and Abstract of Article    & Knowledge \\
Arxiv2023 & 46,198 & 38,863 & 38 & 63/20/17 & Title and Abstract of Paper & Citation \\
Instagram   & 11,339 & 144,010 & 2  & 10/10/90   & Personal Profile of User         & Social \\
Reddit      & 33,434 & 269,442 & 2  & 10/10/90   & Last 3 Posts of User             & Social \\
Ratings     & 24,492 & 186,100 & 5  & 25/25/50   & Name of Product                  & E-commerce \\
Children & 76,875 & 1,162,522 & 24 & 60/20/20 & Name and Description of Book & E-commerce \\
\specialrule{1.5pt}{1.5pt}{1.5pt}
\end{tabular}
\end{table}

\subsection{Baselines and Experimental Settings}
\subsubsection{Baselines.}
\label{sec:baseline_info}
We compare DANCE with representative baselines spanning Classical FL and FGL, Graph condensation, Federated graph condensation, and LLM-based TAG-FGL methods, in order to comprehensively evaluate its effectiveness.

\paragraph{Classical FL and FGL methods.}
\textbf{FedAvg~\cite{fedavg}} is a classical federated learning algorithm that aggregates client models by weighted averaging.
We include it as a standard optimization-based baseline to assess whether performance gains stem from data modeling rather than federated optimization alone.
\textbf{FedSage+~\cite{zhang2021subgraph}} is a subgraph-based federated graph learning method that extends FedSage by explicitly modeling missing neighbors across local subgraphs.
It introduces a neighbor generation mechanism to mitigate structural incompleteness caused by subgraph partitioning, and serves as a canonical structure-aware baseline in subgraph-FL settings.
We include it as a canonical structure-aware baseline in subgraph-FL settings.
\textbf{FedGTA~\cite{li2024fedgta}} is a topology-aware federated graph learning method that introduces topology-aware aggregation to perform personalized model updates across clients.By incorporating local graph structure into the aggregation process through topology-aware smoothing confidence and mixed neighbor features, FedGTA effectively mitigates structural heterogeneity among client subgraphs.
We include it to compare against recent advances in structure-aware subgraph-FL.

\paragraph{Graph condensation methods.}
\textbf{GCond~\cite{jin2021graph}} is a centralized graph condensation method that learns a compact synthetic graph whose training dynamics approximate those of the original graph.It jointly optimizes synthetic node features and graph structure via gradient matching.We include it as a representative condensation baseline to isolate the effectiveness of condensation mechanisms.
\textbf{SFGC~\cite{zheng2023structure}} is a centralized structure-free graph condensation method that compresses graphs into a compact set of synthetic nodes without explicitly constructing a condensed graph topology.
It incorporates structural information implicitly during condensation while avoiding storing or optimizing condensed edges, and is included to contrast with DANCE in terms of whether explicitly modeling condensed structure benefits downstream performance.

\paragraph{Federated graph condensation methods.}
\textbf{FedC4~\cite{fedc4}} is a federated graph condensation framework that compresses local graphs before communication.
We include it as a representative federated condensation baseline to contrast with DANCE in terms of data abstraction and structural modeling.
\textbf{FedGVD~\cite{dai2025fedgvd}} is a generative federated condensation approach that employs server-side virtual nodes and knowledge distillation to share global information.
We include it to compare with data-centric federated condensation methods that do not preserve explicit graph structure.
\textbf{FedGM~\cite{fedgm}} is a data condensation-based federated graph learning framework that addresses data heterogeneity by aggregating condensed graphs as optimization carriers.
We include it as a representative baseline for evaluating how different condensation strategies handle heterogeneity in federated graph learning.

\paragraph{LLM-based TAG-FGL methods.}
\textbf{LLM4RGNN~\cite{zhang2025can}} is an LLM-assisted graph learning framework that leverages the reasoning capabilities of large language models to infer and repair graph structures, particularly under topology perturbations.
By distilling LLM-based structural inference into local models, it enhances robustness to missing or malicious edges without modifying the underlying GNN architecture.
We include it to compare against LLM-based approaches that utilize language models for structure-level reasoning, rather than data condensation or text-aware selection.
\textbf{LLaTA~\cite{zhang2025rethinking}} is an LLM-assisted TAG-FGL framework that integrates large language models into graph structure learning.
It reformulates GSL as a language-aware tree optimization problem and leverages tree-based in-context inference to refine graph topology without parameter tuning.
We include LLaTA as a representative LLM-based baseline that enhances structural modeling in text-attributed graphs, in contrast to condensation-based data selection approaches.
\textbf{LLM4FGL~\cite{yan2025llm4fgl}} is an LLM-driven federated graph learning framework that addresses data heterogeneity by augmenting local graphs through language-model-based neighbor generation and federated reflection.
It decomposes LLM-assisted FGL into data generation and relation inference, enabling data-level augmentation without modifying model architectures.
We include LLM4FGL as a representative LLM-based baseline that tackles heterogeneity at the data level, providing a direct comparison to DANCE from a data-centric perspective.

\subsubsection{Hyper-parameters.}
Table~\ref{tab:hyperparams} summarizes the hyper-parameter settings used in all experiments.
For both clients and the central server, we adopt a 2-layer GCN as the backbone model, with the hidden dimension selected from \{64, 128\}.
Unless otherwise specified, the number of participating clients is fixed to 5.
Federated training is conducted for 200 communication rounds, and in each round, every client performs 3 local training epochs before model aggregation.

For the proposed DANCE framework, we employ a hop-wise neighbor-gated condensation strategy.
The condensation budgets for 1-hop and 2-hop neighbors are set to 3 and 2, respectively.
The summary mixing ratio \textit{mix} is selected from \{0.4, 0.6, 0.8\}.
For the topology reconstruction module, the hyper-parameters $\alpha$ and $\beta$ are searched within \{4, 8, 12\} and \{3, 5, 10\}, respectively.
Across all GNN backbones, the learning rate is set to $1\mathrm{e}{-2}$, the weight decay is set to $5\mathrm{e}{-4}$, and the dropout rate is fixed to 0.5.
All hyper-parameters are selected based on validation performance.
For each experiment, we report the mean and standard deviation over 5 seeds.For textual encoding, we use a frozen pretrained SBERT encoder (all-MiniLM-L6-v2) to extract node-level text representations across all datasets.

\begin{table}[t]
\centering
\caption{Hyper-parameter settings for DANCE.}
\label{tab:hyperparams}
\begin{tabular}{l c}
\toprule
\textbf{Hyper-parameter} & \textbf{Value / Search Range} \\
\midrule
Backbone GNN & 2-layer GCN \\
Hidden dimension & \{64, 128\} \\
Number of clients & 5 \\
Communication rounds & 200 \\
Local epochs per round & 3 \\
\midrule
1-hop budget & 3 \\
2-hop budget & 2 \\
Summary mixing ratio (\textit{mix}) & \{0.4, 0.6, 0.8\} \\
Topology parameter $\alpha$ & \{4, 8, 12\} \\
Topology parameter $\beta$ & \{3, 5, 10\} \\
\midrule
GNN learning rate & $1\mathrm{e}{-2}$ \\
GNN Weight decay & $5\mathrm{e}{-4}$ \\
GNN Dropout rate & 0.5 \\
\bottomrule
\end{tabular}
\end{table}

\section{Algorithm in Pseudo Code}
\begin{algorithm}[t]
\caption{DANCE: Round-wise Condensation and Federated Training}
\label{alg:dance}
\begin{algorithmic}[1]
\REQUIRE Rounds $T$; client fraction $\rho$; budgets $K$, $(B_0,B_1,B_2)$, $B_{\mathrm{tok}}$, $k$.
\STATE Initialize global parameters $\omega^{(0)}=(\theta^{(0)},\phi^{(0)})$.
\FOR{$t=1,\dots,T$}
    \STATE Server samples clients $\mathcal{M}_t$ (rate $\rho$) and broadcasts $\omega^{(t-1)}$.
    \FORALL{$m\in\mathcal{M}_t$ \textbf{in parallel}}
        \STATE \textbf{Text caching:} chunk texts and compute $\{e_{u,r},t_u\}$ via Eq.~\eqref{eq:text_pool} (refresh if encoder changes).
        \STATE \textbf{Node condensation:} compute $(\tilde{y}_v,\pi_v)$ (Eq.~\eqref{eq:pseudo_label}) and $u_v$ (Eq.~\eqref{eq:difficulty});
       cluster per-class embeddings to obtain $\{\mathcal{P}_c\}$; select $\hat{V}^{(m)}_t$ via Eq.~\eqref{eq:topk_select}.
        \STATE \textbf{Text condensation:} build $\tilde{\mathcal{N}}_v^{(2)}$ (Eq.~\eqref{eq:hard_2hop});
       compute neighbor gates $\alpha$ (Eq.~\eqref{eq:neighbor_gating}) and context $c_v$ (Eq.~\eqref{eq:hier_context});
       distill $\tilde{t}_v$ (Eq.~\eqref{eq:agg_text}) and store evidence packs/neighbor cues locally.
        \STATE \textbf{Topology reconstruction:} fuse $X$ (Eq.~\eqref{eq:fused_embed}); update $Z$ (Eq.~\eqref{eq:self_expressive_sparse});
               form $\hat{A}^{(m)}_t$ (Eq.~\eqref{eq:adjacency}).
        \STATE \textbf{Local training \& upload:} train on the condensed TAG using standard subgraph-FL, obtain update $\Delta\omega_m^{(t)}$, and upload $\Delta\omega_m^{(t)}$ only.
        \STATE \textbf{Local audit traces:} summarize evidence packs into keywords/evidence sentences using an on-device LLM; keep all traces local.
    \ENDFOR
    \STATE Server aggregates $\{\Delta\omega_m^{(t)}\}$ to obtain $\omega^{(t)}$.
\ENDFOR
\STATE \textbf{return} $\omega^{(T)}$.
\end{algorithmic}
\end{algorithm}

\begin{algorithm}[H]
\small
\caption{\textbf{Label-aware Node Condensation} (per client $m$ at round $t$)}
\label{alg:node_condense}
\DontPrintSemicolon
\KwIn{Global model $\omega^{(t-1)}$; local TAG $\mathcal{G}_m=(V^{(m)},E^{(m)},S^{(m)})$; labeled set $V_L^{(m)}$ with $\{y_v\}$; threshold $\tau$; prototype cap $P$; node budget $K$ (or ratio $r$)}
\KwOut{Condensed core $\hat{V}^{(m)}_t$; pseudo labels/confidence $\{(\tilde{y}_v,\pi_v)\}$}
\BlankLine
\textbf{Forward (once):} run $\omega^{(t-1)}$ on $\mathcal{G}_m$\;
Obtain embeddings $\{z_v\}_{v\in V^{(m)}}$\;
Obtain predictive distributions $p^{(t-1)}(y\mid v)$\;
\BlankLine
\ForEach{$v\in V^{(m)}$}{
  \eIf{$v\in V_L^{(m)}$}{
    $\tilde{y}_v \leftarrow y_v$\;
  }{
    $\tilde{y}_v \leftarrow \arg\max_{c} p^{(t-1)}(c\mid v)$\;
  }
  $\pi_v \leftarrow \max_{c} p^{(t-1)}(c\mid v)$\;
}
\BlankLine
\ForEach{class $c$}{
  $V_c^{(m)} \leftarrow \{v\in V^{(m)}:\tilde{y}_v=c,\ \pi_v\ge\tau\}$\;
}
\BlankLine
\ForEach{class $c$ with $V_c^{(m)}\neq\emptyset$}{
  $P_c \leftarrow \min(P,\ |V_c^{(m)}|)$\;
  $\mathcal{P}_c \leftarrow \mathrm{Cluster}(\{z_v\}_{v\in V_c^{(m)}};\ P_c)$\;
}
\BlankLine
\ForEach{$v$ with $\pi_v\ge\tau$}{
  $s_v \leftarrow \max_{p\in\mathcal{P}_{\tilde{y}_v}} \kappa(z_v,p)$\;
}
\BlankLine
\If{budget is given as ratio}{
  $K \leftarrow \lceil r\cdot |V^{(m)}|\rceil$\;
}
$N_{\mathrm{conf}} \leftarrow \sum_{c}|V_c^{(m)}|$\;
\ForEach{class $c$}{
  $\bar{K}_c \leftarrow K\cdot \frac{|V_c^{(m)}|}{N_{\mathrm{conf}}}$\;
  $K_c \leftarrow \lfloor \bar{K}_c\rfloor$\;
  $\delta_c \leftarrow \bar{K}_c - K_c$\;
}
$R \leftarrow K - \sum_c K_c$\;
Sort classes by $\delta_c$ in descending order\;
\For{$j=1$ \KwTo $R$}{
  Let $c_j$ be the $j$-th class in the sorted order\;
  $K_{c_j} \leftarrow K_{c_j}+1$\;
}
\BlankLine
$\hat{V}^{(m)}_t \leftarrow \emptyset$\;
\ForEach{class $c$}{
  $\hat{V}^{(m)}_t \leftarrow \hat{V}^{(m)}_t \cup \mathrm{TopK}_{K_c}\big(\{(v,s_v):v\in V_c^{(m)}\}\big)$\;
}
\BlankLine
\textbf{Return} $\hat{V}^{(m)}_t$, $\{(\tilde{y}_v,\pi_v)\}$\;
\end{algorithm}

\begin{algorithm}[H]
\small
\caption{\textbf{Hierarchical Text Condensation} (per client $m$ at round $t$)}
\label{alg:text_condense}
\DontPrintSemicolon
\KwIn{Frozen encoder $\mathrm{Enc}$; chunker $\mathcal{S}(\cdot)$; pooling $\mathrm{Pool}$; (optional) cache $\{e_{u,r},t_u\}$; core $\hat{V}^{(m)}_t$; last-round graph encoder in $\omega^{(t-1)}$; difficulty $\{u_w\}$; hop budgets $(B_0,B_1,B_2)$; token budget $B_{\mathrm{tok}}$; $\Pi_B(\cdot)$; entmax; parameters $(W_q,W_k,W_s,\gamma)$}
\KwOut{Evidence embeddings $\{\tilde{t}_v\}$; neighbor gates $\{\alpha^{(\ell)}\}$; evidence packs $\{\pi\}$ (kept local)}
\BlankLine
\If{cache is missing}{
  \ForEach{$u\in V^{(m)}$}{
    $\{s_{u,r}\}_{r=1}^{R_u} \leftarrow \mathcal{S}(s_u)$\;
    \ForEach{$r\in\{1,\dots,R_u\}$}{
      $e_{u,r} \leftarrow \mathrm{Enc}(s_{u,r})$\;
    }
    $t_u \leftarrow \mathrm{Pool}(\{e_{u,r}\}_{r=1}^{R_u})$\;
  }
}
\textbf{Graph forward (once):} run the last-round graph encoder on the \emph{original} local TAG\;
Obtain graph embeddings $\{g_v\}_{v\in V^{(m)}}$\;
\ForEach{$v\in \hat{V}^{(m)}_t$}{
  $\tilde{\mathcal{N}}_v^{(0)} \leftarrow \{v\}$\;
  $\tilde{\mathcal{N}}_v^{(1)} \leftarrow \mathcal{N}_v^{(1)}$\;
  $\tilde{\mathcal{N}}_v^{(2)} \leftarrow \mathrm{TopK}_{B_2}\big(\{u_w\}_{w\in \mathcal{N}_v^{(2)}}\big)$\;
  \ForEach{$\ell\in\{0,1,2\}$}{
    \ForEach{$u\in \tilde{\mathcal{N}}_v^{(\ell)}$}{
      $s_{v,u} \leftarrow \frac{(W_q g_v)^\top (W_k t_u)}{\sqrt{d}}$\;
    }
    $\tilde{\alpha}^{(\ell)}_{v,\cdot} \leftarrow \mathrm{entmax}(\{s_{v,u}\}_{u\in \tilde{\mathcal{N}}_v^{(\ell)}})$\;
    $\alpha^{(\ell)}_{v,\cdot} \leftarrow \Pi_{B_\ell}(\tilde{\alpha}^{(\ell)}_{v,\cdot})$\;
    $\mathcal{S}_v^{(\ell)} \leftarrow \{u\in \tilde{\mathcal{N}}_v^{(\ell)}:\alpha^{(\ell)}_{v,u}>0\}$\;
  }
  $c_v \leftarrow 0$\;
  \ForEach{$\ell\in\{0,1,2\}$}{
    \ForEach{$u\in\mathcal{S}_v^{(\ell)}$}{
      $c_v \leftarrow c_v + \gamma_\ell\cdot \alpha^{(\ell)}_{v,u}\cdot t_u$\;
    }
  }
  $\mathcal{E}_v \leftarrow \{(u,r): u\in \cup_{\ell}\mathcal{S}_v^{(\ell)},\ r\in[R_u]\}$\;
  $q_v \leftarrow W_s g_v$\;
  \ForEach{$(u,r)\in\mathcal{E}_v$}{
    $a_{v,(u,r)} \leftarrow \frac{q_v^\top e_{u,r}}{\sqrt{d}}$\;
  }
  $\tilde{\pi}_{v,\cdot} \leftarrow \mathrm{entmax}(\{a_{v,(u,r)}\}_{(u,r)\in\mathcal{E}_v})$\;
  $\pi_{v,\cdot} \leftarrow \Pi_{B_{\mathrm{tok}}}(\tilde{\pi}_{v,\cdot})$\;
  $\tilde{t}_v \leftarrow 0$\;
  \ForEach{$(u,r)\in\mathcal{E}_v$}{
    $\tilde{t}_v \leftarrow \tilde{t}_v + \pi_{v,(u,r)}\cdot e_{u,r}$\;
  }
  Store neighbor cues via $\mathrm{TopK}(\alpha^{(\ell)}_{v,\cdot})$\;
  Store evidence pack via $\mathrm{TopK}(\pi_{v,\cdot})$\;
}
\textbf{Return} $\{\tilde{t}_v\}_{v\in\hat{V}^{(m)}_t}$, $\{\alpha^{(\ell)}\}$, $\{\pi\}$\;
\end{algorithm}

\begin{algorithm}[H]
\small
\caption{\textbf{Self-expressive Topology Reconstruction} (per client $m$ at round $t$)}
\label{alg:topo_recon}
\DontPrintSemicolon
\KwIn{Core $\hat{V}^{(m)}_t$; embeddings $\{g_v\}$; evidence $\{\tilde{t}_v\}$; fusion $(W_g,W_t,w)$; evidence prior $S$; candidate size $q$; $(\lambda_1,\lambda_3)$; top-$k$ operator $\mathcal{T}_k$; step size $\eta$; inner iterations $L$}
\KwOut{Reconstructed adjacency $\hat{A}^{(m)}_t$; (optional) coefficients $Z$}
\BlankLine
\ForEach{$v\in \hat{V}^{(m)}_t$}{
  $\alpha_v \leftarrow \sigma\!\big(w^\top[g_v\Vert \tilde{t}_v]\big)$\;
  $x_v \leftarrow \mathrm{LN}\!\big(W_g g_v + \alpha_v\cdot W_t \tilde{t}_v\big)$\;
}
$X \leftarrow [x_v]_{v\in\hat{V}^{(m)}_t}$\;
\BlankLine

\ForEach{$i\in\{1,\dots,K\}$}{
  $\mathcal{C}_x(i) \leftarrow \mathrm{TopK}_q(\{x_i^\top x_j\}_{j\neq i})$\;
  $\mathcal{C}_S(i) \leftarrow \mathrm{TopK}_q(\{S_{ij}\}_{j\neq i})$\;
  $\mathcal{C}(i) \leftarrow \mathcal{C}_x(i)\cup \mathcal{C}_S(i)$\;
}
\BlankLine

Initialize $Z \leftarrow 0$\;
Set $\mathrm{diag}(Z)\leftarrow 0$\;
Mask $Z_{ij}\leftarrow 0$ for all $j\notin\mathcal{C}(i)$\;
\BlankLine

\For{$\ell=1$ \KwTo $L$}{
  $G \leftarrow X^\top(XZ - X)$\;
  Mask $G_{ij}\leftarrow 0$ for all $j\notin\mathcal{C}(i)$\;
  Set $G_{ii}\leftarrow 0$\;
  \BlankLine

  $Z \leftarrow Z - \eta G$\;
  \BlankLine

  \ForEach{masked entry $(i,j)$ with $j\in\mathcal{C}(i)$ and $i\neq j$}{
    $\tau_{ij} \leftarrow \eta\big(\lambda_1 + \lambda_3(1-S_{ij})\big)$\;
    $Z_{ij} \leftarrow \mathrm{sign}(Z_{ij})\cdot \max(|Z_{ij}|-\tau_{ij},\,0)$\;
  }
  \BlankLine

  Set $\mathrm{diag}(Z)\leftarrow 0$\;
  Mask $Z_{ij}\leftarrow 0$ for all $j\notin\mathcal{C}(i)$\;
}
\BlankLine

$W \leftarrow |Z|$\;
$W \leftarrow W + |Z|^\top$\;
$\hat{A}^{(m)}_t \leftarrow \mathcal{T}_k(W)$\;
\BlankLine
\textbf{Return} $\hat{A}^{(m)}_t$ (and $Z$)\;
\end{algorithm}

\section{Proof of Theoretical Analysis}
\label{sec:proof}
\subsection{Budget Guarantees and Cost Decomposition}
\label{sec:budget_cost}

We first formalize the hard budget guarantees induced by the truncation operator $\Pi_B$ used in neighbor gating and chunk selection, and then decompose the per-round client cost into (i) scoring and selection and (ii) budgeted aggregation. 
Throughout, $\Pi_B(\cdot)$ keeps the top-$B$ entries (ties broken by a fixed deterministic rule) and renormalizes.

\begin{assumption}[Bounded representations]
\label{ass:bounded_rep}
There exist constants $G,M>0$ such that along training,
$\|g_v\|_2\le G$ for all graph-side embeddings and $\|e_{u,r}\|_2\le M$ for all chunk embeddings.
This can be enforced in practice via normalization or clipping.
\end{assumption}

\begin{proposition}[Hard budget guarantees]
\label{prop:budget_v3}
Fix a client $m$ and a round $t$. For each core node $v\in\hat{V}_t^{(m)}$:
\begin{enumerate}
\renewcommand{\labelenumi}{(\roman{enumi})}
\item For each hop $\ell\in\{0,1,2\}$, the selected neighbor set satisfies $|\mathcal{S}_v^{(\ell)}|\le B_\ell$.
\item The chunk-selection weights $\pi_{v,\cdot}$ have at most $B_{\mathrm{tok}}$ nonzeros, hence at most $B_{\mathrm{tok}}$ chunks contribute to $\tilde{t}_v$.
\item Across the condensed core, the total number of nonzero chunk contributions to $\{\tilde{t}_v\}_{v\in\hat{V}_t^{(m)}}$ is at most $K\cdot B_{\mathrm{tok}}$ per round.
\end{enumerate}
\end{proposition}

\begin{proof}
(i) In Eq.~\eqref{eq:neighbor_gating}, $\alpha^{(\ell)}_{v,\cdot}=\Pi_{B_\ell}(\cdot)$ keeps at most $B_\ell$ nonzeros, hence $|\mathcal{S}_v^{(\ell)}|\le B_\ell$.

(ii) In Eq.~\eqref{eq:chunk_similarity}, $\pi_{v,\cdot}=\Pi_{B_{\mathrm{tok}}}(\cdot)$ keeps at most $B_{\mathrm{tok}}$ nonzeros.

(iii) By (ii), each core node aggregates at most $B_{\mathrm{tok}}$ chunk embeddings in Eq.~\eqref{eq:agg_text}. Summing over $K=|\hat{V}_t^{(m)}|$ yields at most $K B_{\mathrm{tok}}$ nonzero chunk contributions per round.
\end{proof}

\begin{corollary}[Per-round scoring vs.\ aggregation cost]
\label{cor:cost_v3}
Assume each node text is split into at most $R_{\max}$ chunks, i.e., $R_u\le R_{\max}$.
For a core node $v$, the neighbor scoring in Eq.~\eqref{eq:neighbor_score} costs
\[
O\!\left(d\sum_{\ell\in\{0,1,2\}}|\tilde{\mathcal{N}}_v^{(\ell)}|\right),
\]
while the \emph{budgeted neighbor aggregation} in Eq.~\eqref{eq:hier_context} costs
\[
O\!\left(d\sum_{\ell\in\{0,1,2\}} B_\ell\right).
\]
Moreover, chunk candidates are collected only from selected neighbors, so
\[
|\mathcal{E}_v|
\le
\Big(\sum_{\ell}|\mathcal{S}_v^{(\ell)}|\Big)\,R_{\max}
\le
\Big(\sum_{\ell}B_\ell\Big)\,R_{\max}.
\]
Therefore, chunk scoring in Eq.~\eqref{eq:chunk_similarity} costs
\[
O\!\left(d\Big(\sum_{\ell}B_\ell\Big)R_{\max}\right),
\]
while the \emph{budgeted chunk aggregation} in Eq.~\eqref{eq:agg_text} costs
\[
O(d\,B_{\mathrm{tok}}).
\]
Summing over $K$ core nodes yields the total per-round cost of evidence selection/aggregation on client $m$.
In addition, the graph encoder forward pass on the original local subgraph contributes an additive backbone cost $C_{\mathrm{GNN}}(G_m)$ per round.
\end{corollary}

\begin{proof}
Each score in Eq.~\eqref{eq:neighbor_score} or Eq.~\eqref{eq:chunk_similarity} is an inner product in $\mathbb{R}^d$, hence $O(d)$ time.
The number of neighbor scores equals $|\tilde{\mathcal{N}}_v^{(\ell)}|$ per hop, giving the stated neighbor scoring cost.

By Proposition~\ref{prop:budget_v3}(i), at most $B_\ell$ neighbors per hop contribute to Eq.~\eqref{eq:hier_context}$,$ and each contribution is a $d$-dimensional weighted sum, giving
$O(d\sum_\ell B_\ell)$.

For chunks, $\mathcal{E}_v$ enumerates chunks only from selected neighbors.
Each selected neighbor contributes at most $R_{\max}$ chunks, hence
$|\mathcal{E}_v|\le (\sum_\ell B_\ell)R_{\max}$, and chunk scoring costs
$O(d(\sum_\ell B_\ell)R_{\max})$.
Finally, Proposition~\ref{prop:budget_v3}(ii) ensures at most $B_{\mathrm{tok}}$ chunks contribute to Eq.~\eqref{eq:agg_text}, yielding aggregation cost $O(dB_{\mathrm{tok}})$.
Summing over $K$ core nodes gives the per-round client cost, and the backbone term $C_{\mathrm{GNN}}(G_m)$ follows from the required forward pass producing $\{g_v\}$.
\end{proof}

The above bounds capture the dominant arithmetic cost from inner products and budgeted weighted sums.
Selection operators such as $\mathrm{entmax}(\cdot)$ and top-$B$ truncation can incur additional $O(n\log n)$ (or $O(n\log B)$) time for sorting/thresholding over $n$ candidates; these terms are standard and omitted from the main expressions for clarity.

\subsection{Approximation Error from Hard Budget Projection}
\label{sec:approx_err}

We quantify the approximation error introduced solely by the hard truncation $\Pi_{B_{\mathrm{tok}}}$ in Eq.~\eqref{eq:chunk_similarity}, relative to using the \emph{same} scoring rule without truncation (i.e., only removing the top-$B_{\mathrm{tok}}$ renormalization step).

\begin{definition}[Tail mass under top-$B$ truncation]
\label{def:tailmass_v3}
For a probability vector $p\in\Delta^n$, let $T_B(p)$ denote the indices of its top-$B$ entries (ties broken by a fixed rule), and define the tail mass
\[
\delta_B(p)\triangleq \sum_{i\notin T_B(p)} p_i.
\]
\end{definition}

\begin{lemma}[Truncation bound for probability-weighted sums]
\label{lem:trunc_v3}
Let $\{e_i\}_{i=1}^n\subset\mathbb{R}^d$ satisfy $\|e_i\|_2\le M$.
For any $p\in\Delta^n$ and $\tilde{p}=\Pi_B(p)$,
\[
\left\|\sum_{i=1}^n p_i e_i \;-\; \sum_{i=1}^n \tilde{p}_i e_i\right\|_2
\;\le\;
2M\cdot \delta_B(p).
\]
\end{lemma}

\begin{proof}
Let $T=T_B(p)$ and $s=\sum_{i\in T}p_i=1-\delta_B(p)$.
By definition of $\Pi_B$, $\tilde{p}_i=\frac{p_i}{s}$ for $i\in T$ and $\tilde{p}_i=0$ otherwise, hence
$\sum_{i=1}^n \tilde{p}_i e_i=\sum_{i\in T}\frac{p_i}{s}e_i$.
Decompose the difference into discarded mass and renormalization bias:
\[
\sum_{i=1}^n p_i e_i-\sum_{i\in T}\frac{p_i}{s}e_i
=
\underbrace{\sum_{i\notin T}p_i e_i}_{(\mathrm{A})}
+\underbrace{\sum_{i\in T}p_i\left(1-\frac{1}{s}\right)e_i}_{(\mathrm{B})}.
\]
For term (A), $\|(\mathrm{A})\|_2\le \sum_{i\notin T}p_i\|e_i\|_2\le M\,\delta_B(p)$.
For term (B), note that $\left|1-\frac{1}{s}\right|=\frac{1-s}{s}=\frac{\delta_B(p)}{1-\delta_B(p)}\le \frac{\delta_B(p)}{1-\delta_B(p)}$ and
\[
\|(\mathrm{B})\|_2
\le
\left(\frac{1}{s}-1\right)\sum_{i\in T}p_i\|e_i\|_2
\le
\left(\frac{1}{s}-1\right) M s
=
M(1-s)
=
M\,\delta_B(p).
\]
Combining both bounds yields the result.
\end{proof}

\begin{proposition}[Chunk condensation approximation error in DANCE]
\label{prop:chunk_err_v3}
Fix a core node $v$ and its candidate chunk set $\mathcal{E}_v$.
Let
\[
p_v \triangleq \mathrm{entmax}\!\left(\{a_{v,(u,r)}\}_{(u,r)\in\mathcal{E}_v}\right)\in\Delta^{|\mathcal{E}_v|}
\]
be the (untruncated) probability vector produced by entmax, and let
$\pi_v=\Pi_{B_{\mathrm{tok}}}(p_v)$ be the budgeted weights in Eq.~\eqref{eq:chunk_similarity}.
Define the unbudgeted and budgeted evidence embeddings
\[
t_v^{\mathrm{full}} \triangleq \sum_{(u,r)\in\mathcal{E}_v} (p_v)_{(u,r)}\, e_{u,r},
\qquad
\tilde{t}_v \triangleq \sum_{(u,r)\in\mathcal{E}_v} (\pi_v)_{(u,r)}\, e_{u,r}.
\]
Under Assumption~\ref{ass:bounded_rep},
\[
\|\tilde{t}_v-t_v^{\mathrm{full}}\|_2 \;\le\; 2M\cdot \delta_{B_{\mathrm{tok}}}(p_v).
\]
An analogous bound holds for neighbor gating at hop $\ell$ by replacing $p_v$ with the corresponding (untruncated) neighbor-weight vector and replacing $B_{\mathrm{tok}}$ with $B_\ell$.
\end{proposition}

\begin{proof}
Apply Lemma~\ref{lem:trunc_v3} to the collection $\{e_{u,r}\}_{(u,r)\in\mathcal{E}_v}$ with $p=p_v$ and $\tilde{p}=\Pi_{B_{\mathrm{tok}}}(p_v)=\pi_v$.
\end{proof}

\noindent\textbf{Remark.}
The error scales with the discarded probability mass $\delta_{B_{\mathrm{tok}}}(p_v)$.
When entmax already produces a sparse distribution (small tail mass), the additional hard truncation incurs a correspondingly small approximation error.

\subsection{Stability of Model-in-the-loop Evidence Refresh}
\label{sec:stability}

Finally, we characterize when the round-wise refresh changes the selected neighbors/chunks.
DANCE forms selection weights by applying $\mathrm{entmax}(\cdot)$ to the scores and then performing hard top-$B$ truncation via $\Pi_B$.
Since both steps preserve the score ordering (up to ties), the selected index set is stable as long as the score ranking around the top-$B$ boundary does not change.

\begin{definition}[Top-$B$ index set]
\label{def:topB_set}
For a vector $r\in\mathbb{R}^n$, let $T_B(r)$ denote the indices of its top-$B$ entries (ties broken by a fixed rule).
\end{definition}

\begin{lemma}[Order preservation implies identical top-$B$ sets]
\label{lem:order_preserve_v3}
Let $\varphi:\mathbb{R}\to\mathbb{R}$ be strictly increasing and apply it elementwise to $r\in\mathbb{R}^n$.
Then for any $B$, $T_B(r)=T_B(\varphi(r))$.
\end{lemma}

\begin{proof}
For any $i,j$, strict monotonicity gives $r_i>r_j \Leftrightarrow \varphi(r_i)>\varphi(r_j)$, so the ordering (and hence the top-$B$ set) is unchanged.
\end{proof}

\begin{assumption}[Local Lipschitzness of scores along the trajectory]
\label{ass:lipschitz_score_v3}
Along the parameter trajectory $\{\omega^{(t)}\}_{t=0}^T$, for any fixed node $v$ and any candidate index $j$ (neighbor or chunk),
the corresponding score function satisfies
\[
\big|\mathrm{score}_j(\omega)-\mathrm{score}_j(\omega')\big|
\le L_s\|\omega-\omega'\|_2
\quad\text{for all }\omega,\omega'\text{ on the trajectory}.
\]
A sufficient condition is $\|\nabla_\omega \mathrm{score}_j(\omega)\|_2\le L_s$ along the trajectory (e.g., via weight decay/clipping and Assumption~\ref{ass:bounded_rep}).
\end{assumption}

\begin{definition}[Top-$B$ margin]
\label{def:margin_v3}
Let $r(\omega)\in\mathbb{R}^n$ be candidate scores sorted as
$r_{(1)}(\omega)\ge\cdots\ge r_{(n)}(\omega)$.
Define the margin
\[
\Delta_B(\omega)\triangleq r_{(B)}(\omega)-r_{(B+1)}(\omega).
\]
\end{definition}

\begin{theorem}[Stability of DANCE evidence selection across rounds]
\label{thm:stability_v3}
Fix a node $v$ and consider either neighbor selection at hop $\ell$ (budget $B_\ell$) or chunk selection (budget $B_{\mathrm{tok}}$).
Let $r(\omega)$ denote the corresponding \emph{pre-entmax} score vector (i.e., $s_{v,u}$ or $a_{v,(u,r)}$).
Under Assumption~\ref{ass:lipschitz_score_v3}, if
\[
\|\omega-\omega'\|_2 \le \frac{\Delta_B(\omega)}{2L_s},
\]
then the selected index set after entmax and truncation is invariant:
\[
T_B\!\Big(\Pi_B(\mathrm{entmax}(r(\omega)))\Big)
=
T_B\!\Big(\Pi_B(\mathrm{entmax}(r(\omega')))\Big),
\]
where $B$ is $B_\ell$ or $B_{\mathrm{tok}}$ accordingly.
\end{theorem}

\begin{proof}
By Assumption~\ref{ass:lipschitz_score_v3}, for every coordinate $j$,
$|r_j(\omega)-r_j(\omega')|\le L_s\|\omega-\omega'\|_2$.
If $\|\omega-\omega'\|_2 \le \Delta_B(\omega)/(2L_s)$, then the standard ranking-stability argument implies
$T_B(r(\omega))=T_B(r(\omega'))$, since no element below rank $B$ can overtake the $B$-th element (and vice versa) under such bounded perturbations.

Next, $\mathrm{entmax}(\cdot)$ is applied elementwise through a monotone transformation of scores prior to normalization, and thus preserves the ordering induced by $r(\cdot)$; by Lemma~\ref{lem:order_preserve_v3},
$T_B(r(\omega))=T_B(\mathrm{entmax}(r(\omega)))$ and similarly for $\omega'$.

Finally, $\Pi_B(\cdot)$ keeps exactly the top-$B$ indices of its input by definition, so applying $\Pi_B$ does not change the top-$B$ index set.
Combining the above yields the claim.
\end{proof}

Theorem~\ref{thm:stability_v3} formalizes \emph{controlled adaptivity} of model-in-the-loop refresh:
the selected neighbors/chunks remain stable under small global-model drift when the top-$B$ margin is non-trivial, and updates occur only when the evolving model induces sufficient score reordering.

\end{document}